\documentclass[conference]{IEEEtran}
\usepackage{amsmath}
\usepackage{amsfonts}
\usepackage{algorithm}
\usepackage{algorithmic}
\usepackage{multirow}
\usepackage{color}
\usepackage{graphicx}
\usepackage{subfigure}
\usepackage{float}
\usepackage{booktabs}
\usepackage[colorlinks]{hyperref}
\usepackage{bm}
\AtBeginDocument{%
  \providecommand\BibTeX{{%
    \normalfont B\kern-0.5em{\scshape i\kern-0.25em b}\kern-0.8em\TeX}}}

\newtheorem{definition}{Definition}

\usepackage{cite}

\ifCLASSINFOpdf
\else
\fi

\begin{document}
%
\title{Learning to Optimize Industry-Scale \\Dynamic Pickup and Delivery Problems}


\author{\IEEEauthorblockN{
Xijun Li\IEEEauthorrefmark{3}\IEEEauthorrefmark{5}\IEEEauthorrefmark{1},
Weilin Luo\IEEEauthorrefmark{5}\IEEEauthorrefmark{4}\IEEEauthorrefmark{1},
Mingxuan Yuan\IEEEauthorrefmark{5},
Jun Wang\IEEEauthorrefmark{5}\IEEEauthorrefmark{2}, 
Jiawen Lu\IEEEauthorrefmark{5}, 
Jie Wang\IEEEauthorrefmark{3}, 
Jinhu L{\"u}\IEEEauthorrefmark{4} 
and Jia Zeng\IEEEauthorrefmark{5}
}
\IEEEauthorblockA{
\IEEEauthorrefmark{3}MIRA Lab, USTC\\
\IEEEauthorrefmark{5}Noah's Ark Lab, Huawei \\
\IEEEauthorrefmark{2}University College London\\
\IEEEauthorrefmark{4}Beihang University\\
\IEEEauthorrefmark{1}Equal contributions}
}


%


\maketitle

\begin{abstract}
The Dynamic Pickup and Delivery Problem (DPDP) is aimed at dynamically scheduling vehicles among multiple sites in order to minimize the cost when delivery orders are not known a priori. Although DPDP plays an important role in modern logistics and supply chain management, state-of-the-art DPDP algorithms are still limited on their solution quality and efficiency. In practice, they fail to provide a scalable solution as the numbers of vehicles and sites become large. In this paper, we propose a data-driven approach, \underline{S}patial-\underline{T}emporal Aided \underline{D}ouble \underline{D}eep \underline{G}raph \underline{N}etwork (ST-DDGN), to solve industry-scale DPDP. In our method, the delivery demands are first forecast using spatial-temporal prediction method, which guides the neural network to perceive spatial-temporal distribution of delivery demand when dispatching vehicles. Besides, the relationships of individuals such as vehicles are modelled by establishing a graph-based value function. ST-DDGN incorporates attention-based graph embedding with Double DQN (DDQN). As such, it can make the inference across vehicles more efficiently compared with traditional methods. Our method is entirely data driven and thus adaptive, i.e., the relational representation of adjacent vehicles can be learned and corrected by ST-DDGN from data periodically. We have conducted extensive experiments over real-world data to evaluate our solution. The results show that ST-DDGN reduces 11.27\% number of the used vehicles and decreases 13.12\% total transportation cost on average over the strong baselines, including the heuristic algorithm deployed in our UAT (User Acceptance Test) environment and a variety of vanilla DRL methods. We are due to fully deploy our solution into our online logistics system and it is estimated that millions of USD logistics cost can be saved per year.
\end{abstract}


%
\IEEEpeerreviewmaketitle


\section{Introduction}
\label{sec: introduction}

The Dynamic Pickup and Delivery Problem (DPDP) is a fundamental problem in manufacturing enterprises. Huawei, as one of the largest global communication device vendors, manufactures billions of productions in hundreds of factories every year. A large amount of cargoes (including the materials, productions and semi-finished productions) need to be delivered among factories and warehouses at multiple sites/locations\footnote{Note that sites, locations and nodes are used to represent factories/warehouse, which are interchangeably used in the rest of paper.} during the manufacturing. Due to the uncertainty of customer requirements and producing processes, most of delivery requirements cannot be fully decided beforehand. The delivery orders, including the pickup sites, delivery sites, the amount of cargoes and the time requirement, dynamically come and a fleet of homogeneous vehicles is periodically scheduled to serve these orders. Due to the large amount of delivery requests, even a small improvement of the logistics efficiency can bring significant benefits. Therefore, a good dispatching plan must take into consideration both the level of service (LOS) and the logistics cost.


Technically, DPDP is an enhanced variant of traditional Vehicle Routing Problem (VRP) which is one of the most widely studied combinatorial optimization problems. VRP aims to find the optimal set of routes for a fleet of vehicles to traverse, in order to satisfy the demands of customers distributed at different sites. It has been proven that VRP belongs to the NP-Hard problem~\cite{cordeau2008recent}. Even without the consideration of dynamism, the Pickup and Delivery Problem (PDP)~\cite{cordeau2008recent} is hard to solve, which has additional complexities over vanilla VRP. Due to the high complexities, the majority of proposed solutions for PDP are heuristic algorithms. Only a few exact solutions~\cite{lu2004exact,mahmoudi2016finding} are proposed to PDP. When PDP becomes dynamic, the exact solution is intractable except for specific scenarios. Similarly, solutions for DPDP also fall into two categories. The first one~\cite{mitrovic2004waiting, gendreau2006neighborhood, pureza2008waiting} is heuristic where the basic idea is to insert the new delivery order into the planned routes. The other is an operational research-based optimization method~\cite{savelsbergh1998drive}, in which the dynamic problem is decomposed into a series of static problems with a subset of known delivery orders and each static problem is optimized by linear programming or mixed integer programming.


There are several reasons~\cite{li2019cooperative} as to why traditional methods fail in solving DPDP or related dynamic optimization problems. They are summarized as follows. First, it is hard to take into account future orders in multi-stage static optimization~\cite{savelsbergh1998drive} or insertion heuristics~\cite{mitrovic2004waiting, gendreau2006neighborhood, pureza2008waiting} due to the high uncertainty of future orders. Multiple external dynamic factors, such as special days/events, market fluctuations, unstable policies, \textit{etc.}, all bring high uncertainty to the future orders. Second, such uncertainty might be escalated by the interaction between the current optimized solution obtained by traditional methods and the future orders. Third, not all complex business rules can be formulated in the constraints of linear or convex forms, which hence makes it difficult to model and solve the dynamic problem using operational research-based optimization methods such as linear programming and convex optimization. The operational research-based methods can only be applied to the problems with specific structures when problem scale becomes large. Also, the heuristic algorithms may fall into sub-optimal solutions because of its `myopic' view.

In recent years, people began to use Reinforcement Learning (RL) to improve the `myopic view' of heuristic algorithms. Previous works~\cite{khalil2017learning, kool2018attention, balaji2019orl} show that for Traveling Salesman Problem (TSP), VRP, and PDP, RL techniques can produce competitive results and perform better under dynamic and stochastic circumstances. Thus, due to its data driven nature, RL has the potential of improving the solution of more complex DPDPs. In this paper, we propose a \textit{route-centric Markov} Decision Process (MDP) for DPDP in our scenario. And an
end-to-end RL solution, \underline{S}patial-\underline{T}emporal Aided \underline{D}ouble \underline{D}eep \underline{G}raph \underline{N}etwork (ST-DDGN), is presented to solve the problem. This solution incorporates attention-based graph embedding with Double DQN (DDQN), to coordinate the cooperation and competition among vehicles in a dynamic environment. Besides, during the learning process, it is guided by spatial-temporal prediction of delivery demand. Extensive experiments over real-world data collected from a manufacturing campus with 27 factories verified that the proposed approach performs much better than the existing strong baselines. 

Our contribution is four-fold: 

\begin{itemize}
	
	\item We provide a unified account for all the practical enterprise-scale constraints, including the time window, capacity, last-in-first-out and back-to-depot constraints together whereas previous approaches~\cite{mitrovic2004waiting, pureza2008waiting, cordeau2008recent, lu2004exact, mahmoudi2016finding} are dedicated only to some of them. 
	
	\item We propose the first graph-based relational learning approach for the industry-scale DPDP, where the relation between vehicles in a dynamic dispatching environment is modelled by attention-based graph convolution. 
	
	\item The proposed approach is aided by spatial-temporal prediction of delivery demand so as to accelerate the policy learning. Meanwhile, we propose the constraint embedding technique to further decrease inference time.
	
	\item Extensive experiments have shown the large improvements in practice over real-world data. Our approach reduces 11.27\% number of the used vehicles and decreases 13.12\% total transportation cost on average over the strong baselines.
	
\end{itemize}

The remainder of this paper is organized as follows. Related works are discussed in Section~\ref{sec: related work}. We define our problem in Section~\ref{sec: problem} and present our solution in Section~\ref{sec: solution}. The experimental results are reported in Section~\ref{sec: experiment} followed by supplementary materials in Section~\ref{sec: supplementary materials}. The work is concluded in Section~\ref{sec: conclusion}.


\section{Related Work}
\label{sec: related work}

\subsection{The dynamic pickup and delivery problem}
DPDP extensively exists in real world, such as courier operations and door-to-door transportation services~\cite{berbeglia2010dynamic}. Unlike the classical VRP, there are much less studies on DPDP owing to its greater complexity. The solutions for DPDP on literature can be divided into two categories. The first one, which is straightforward, is to adapt a traditional optimization algorithm that solves the static version of the problem (i.e., PDP). Savelsbergh \textit{et al}.~\cite{savelsbergh1998drive} firstly proposed an optimization based algorithm where the problem is decomposed into a series of static problems with a subset of known delivery order over a rolling-horizon framework. The static problem is solved by using a branch-and-price heuristics. One major drawback of this method is high complexity of computation because it needs to optimize each static subproblem. Therefore it is inadequate for real-time requirement of our company scenario.

The second one, which is widely used, is a hybridization of heuristic and local search. Gendreau \textit{et al}.~\cite{gendreau2006neighborhood} designed a tabu search algorithm and used a neighborhood structure based on ejection chains. Similar to the problem decomposition in~\cite{savelsbergh1998drive}, Minic \textit{et al}.~\cite{mitrovic2004waiting} proposed a two-phased heuristic to optimize each static subproblem, where the first phase is a cheapest insertion heuristic and the second one is a tabu search algorithm analogous to~\cite{gendreau2006neighborhood}. Saez \textit{et al}.~\cite{saez2008hybrid} developed an adaptive predictive control algorithm based on fuzzy clustering and genetic algorithms. Moreover, Pureza \textit{et al}.~\cite{pureza2008waiting} showed that incorporating waiting and buffering strategy to the above heuristics would lead to less number of used vehicles or total travel distance. The heuristic methods are normally very efficient to support real-time calculation. However, they may fall into sub-optimal results due to the lacking of global information. The algorithmic accuracy can be further improved if we can improve the `myopic view' problem.

\subsection{Reinforcement learning for combinatorial optimization}
Some previous works~\cite{khalil2017learning, kool2018attention, balaji2019orl} demonstrate that reinforcement learning techniques might be a promising structure to improve the `myopic view' of heuristic algorithms in combinatorial optimization. Moreover, relational learning plays significant role in terms of blending generalization power with reinforcement learning~\cite{zambaldi2018relational}. The underlying relation can be captured in multiple ways. For a given agent, some approaches~\cite{sukhbaatar2016learning, peng2017multiagent} explicitly construct a relation kernel to observe other agents' states. While other approaches~\cite{yang2018mean, jaques2018social} implicitly represent the relation by referring other agents' rewards or actions. Recently, graph convolutional network~\cite{niepert2016learning} is widely applied in Deep RL community to reveal the hidden correlation of entities, actions and objects. Zambaldi \textit{et al}.~\cite{zambaldi2018relational} construct a graph convolutional network with multi-head dot-product attention to reason the relation between entities. Based on~\cite{zambaldi2018relational}, Jiang \textit{et al}.~\cite{jiang2018graph} extend this framework into multi-agent RL, learning the relation representation between agents. In additions, Malysheva \textit{et al}.~\cite{malysheva2018deep} adopt a relevance graph to learn the relation representation with an heuristic-based loss function. Li \textit{et al}.~\cite{li2019towards} present an actor-critic framework based on graph neural network to obtain graph-based representation, which can be used for learning complex tasks from a curriculum. 
In this paper, we present the first graph relational learning approach in the context of DPDP. Our method provides a unified solution for large-scale practical DPDP by integrating the flexibility of constraint embedding and efficiency of graph network inference among vehicles.


\vspace{-1ex}

\section{The Problem}
\label{sec: problem}

\begin{figure}[t]
	\centering
	\includegraphics[width=1.0\linewidth]{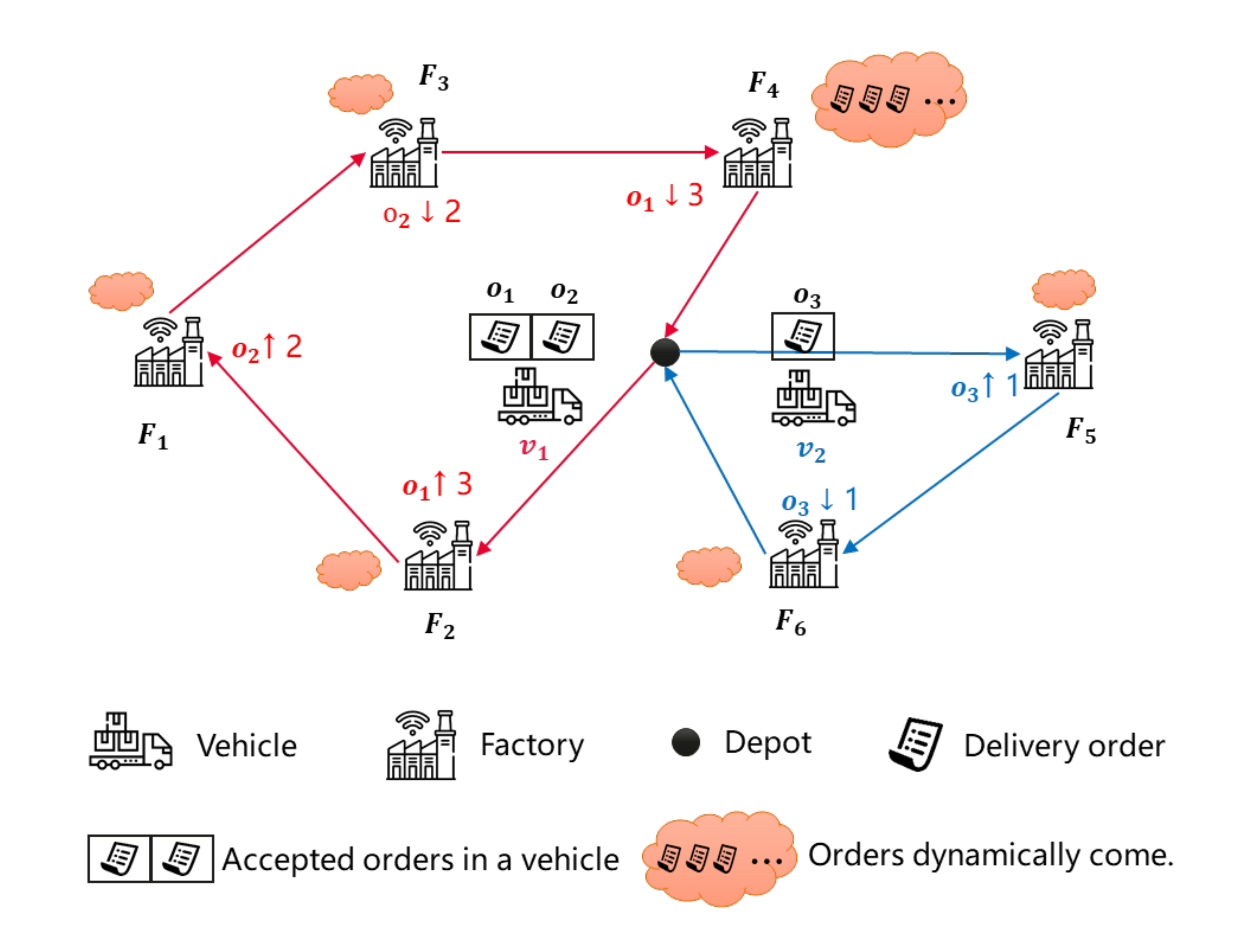}
	\caption{A toy example of DPDP: 1) Each vehicle starts from a depot, serves delivery orders (`$\uparrow$' and `$\downarrow$' represent loading and unloading cargoes respectively and the Last In, First Out (LIFO) constraint must be satisfied), and goes back to the depot; 2) Delivery orders are generated dynamically in various factories, which specifies the pickup and delivery nodes, amount of cargoes to be delivered, \textit{etc.}; 3) Each vehicle is associated with an order list and  a route (dynamically change). The order list contains the delivery orders accepted by the vehicle. And the route is the sequence of traversed nodes by the vehicle, which is arranged according to the order list.}
	\label{fig:toy example}
\end{figure}

In this section, we describe the DPDP from our practical logistics scenario. To intuitively demonstrate the problem, we illustrate a toy example in Fig.~\ref{fig:toy example}. A delivery order specifies the pickup node, amount of cargoes to be picked, delivery node and time window requirement. Within a day, there are multiple factories dynamically generating delivery orders. For this problem, our goal is to generate a dispatching plan for a fleet of vehicles to fulfil all the delivery orders with minimum transportation cost. The dispatching plan consists of two coupled parts: order assignment and route planning.


The inputs of the problem include:

1. A road network $G=(N,A)$, which is a complete directed graph. $N$ is the node set which consists of multiple depots $\{w_i|i=1,...,m\}$ and geographically distributed factories $\{F_i|i=1,...,n\}$. $A=\{<i,j>|i,j\in N\}$ is the arc set. Each arc $<i,j>$ is associated with a non-negative transportation distance $d_{i,j}$ from node $i$ to node $j$.
 
2. A delivery order $o^i=(F^i_p, F^i_d, q^i, t^i_c, t^i_l)$, where $F^i_p, F^i_d\in N$ respectively represent the pickup and delivery node of the order; $q_i$ refers to the amount of cargoes to be delivered; $t^i_c$ is the creation time of the order (also the earliest time) that a vehicle can pickup cargoes at the pickup node, and $t^i_l$ is the latest time for a vehicle to deliver these cargoes to the delivery node. It is worth noting that within a day, the delivery orders are dynamically generated in different factories at any time.

3. A fleet of homogeneous vehicles $V=\{k|k= 1,2,..,K\}$. Each vehicle $k$ is associated with a configuration $conf_k=(w_k, Q, \mu, \delta)$, where $w_k$ refers to the starting depot of vehicle $k$; $Q$ is the maximum loading capacity; $\mu$ is the fixed cost of using a vehicle (usually in practical scenario $\mu$ is considerably large relatively to operation cost); and $\delta$ denotes the operation cost of a vehicle per kilometer such as fuel cost, maintenance fee, wage payment, \textit{etc.} Note that a vehicle is either waiting at a factory or in-service (i.e. on the way to serve delivery orders). Both $\mu$ and $\delta$ are counted in total transportation cost.

The outputs of the dynamic problem are:

1. Order assignment $OA=\{(o^i, k)\}$, which consists of multiple matches between vehicles and orders. $(o^i, k)$ represents dispatching vehicle $k$ to serve order $o^i$.

2. Route planning $RP=\{rp_k|k=1,...,K\}$ for the fleet. A route $rp_k$ of vehicle $k$ specifies the traverse sequence of the vehicle according to its accepted delivery orders. For example, in Fig.~\ref{fig:toy example}, vehicle $v_1$ has accepted two orders $o_1$ and $o_2$. Then its one possible route is $w$ (the depot) $\rightarrow F_2 \rightarrow F_1 \rightarrow F_3 \rightarrow F_4 \rightarrow w$. 

In addition to common constraints of the classic DPDP~\cite{cordeau2008recent}, our problem is also subject to the following: 1) Time window constraint. Each order $o_i$ is supposed to be severed within its earliest and latest service time; 2) Last In, First Out (LIFO) loading constraint. Every cargo must be loaded and unloaded according to LIFO principle; 3) Capacity constraint. For each vehicle $k$, the total capacity of loaded cargoes cannot exceed the maximum loading capacity $Q$; 4) Back-to-depot constraint. For each vehicle $k$, it should start from the depot $w_k$, serve all orders it has accepted, and finally go back to the depot.

Besides the above constraints, we also consider the following important fact in practice:
No interference with in-service vehicle. It is not allowed to change the destination of a vehicle when the vehicle is in-service. However we can adjust its route when the vehicle waits at factories or depots.


For DPDP from our scenario, the objective is to minimize the total transportation cost for serving all delivery orders.


\vspace{-1ex}
\section{The Solution}

\textcolor{black}{We present our solution to DPDP in this section, which works as follows. We process the delivery orders in the ascending order of creation time. In other words, a delivery order is immediately processed without being postponed. And then the order and associated information (such as spatial-temporal demand prediction and real-time vehicle geographical location) is together sent to a relational learning framework that decides which vehicle to serve the order.}
\label{sec: solution}


\vspace{-1ex}

\subsection{Introduction of spatial-temporal distribution}
\label{subsec: st_distribution}

\subsubsection{Delivery demand prediction}
\begin{figure}[t]
	\centering
	\includegraphics[width=0.9\linewidth]{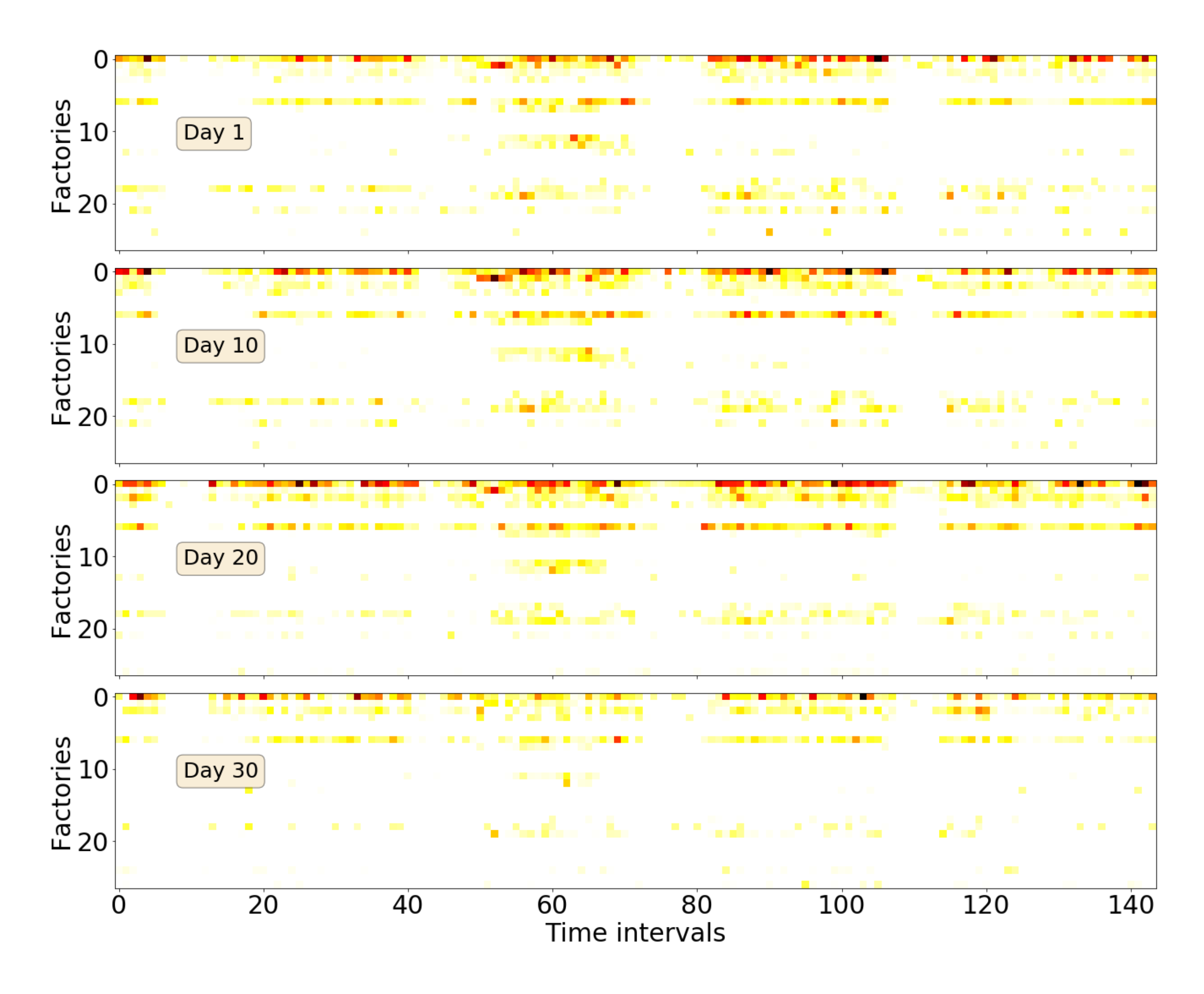}
	\caption {Spatial-temporal distribution of delivery demand of four different days. The four days' data is selected from the same month. Each day is split into $144$ time intervals of equal duration. And there are total $27$ factories in our manufacturing campus. Thus, there are $27 \times 144$ grids in each subfigure. The darker a grid is, the stronger the delivery demand of the grid is.}
	\label{fig: order_distribution}
\end{figure}

Due to multiple factors such as the market cycle, weather, rhythm of production, \textit{etc.}, there is a strong pattern in the spatial-temporal distribution of delivery demand. We first attempt to reveal the pattern. The delivery order data of four days are selected from the same month. Each day is split into $144$ equal-duration intervals (i.e., each interval is with a duration of $10$ minutes). For each factory and each time interval, we sum up the amount of cargoes in delivery orders associated with the factory and time interval, to get the accumulated delivery demand.
\textcolor{black}{In this way, we could construct a spatial-temporal distribution of delivery demand, which is visualized in Fig.~\ref{fig: order_distribution}. In this figure, the vertical axis represents $27$ factories in our manufacturing campus and horizontal axis denotes the $144$ time intervals. Firstly we find that the patterns of four day are similar to some extent; the closer the day, the more similar their patterns. Secondly, from the spatial perspective, it is obvious that some factories has much higher delivery demand than other factories on all the four days. Moreover, from the temporal perspective, for a specific factory, the delivery demand concentrates on several time intervals, such as 10 a.m. to 12 a.m. and 2 p.m. to 5 p.m. Based on above observations, it is believed that we can predict the Spatial-Temporal Distribution (STD) of delivery demand using historical data, to help solve DPDP. Formally, we give the definition of STD matrix of delivery demand and then describe how to predict it.}
\vspace{-3ex}

\textcolor{black}{
\begin{definition}[STD Matrix]
\label{def: std matrix}
$\bf{E}^d$ $=[e^d_{i,j}]_{n\times T}\in\mathbb{R}^{n\times T}$ represents the spatial-temporal distribution of delivery demand of day $d$, where $n$ denotes the number of factories and $T$ is the number of equal-duration time intervals within day $d$. Each element $e^d_{i,j}$ represents the total amount of cargoes (to be delivered) at factory $F_i$ and within time interval $TI_j$, which is calculated by:
\begin{equation}
    O_{i,j} = \{o^h | F_p^h=F_i, t_c^h \in TI_j\}
\end{equation}
\begin{equation}
    e^d_{i,j} = \sum_{o^h \in O_{i,j}} q^h
\end{equation}
where $h$ is the index of delivery orders and $O_{i,j}$ is the set of delivery orders associated with factory $F_i$ and time interval $TI_j$. The time interval is a left-closed and right-open interval and its duration is dependent on $T$.
\end{definition}
}


\textcolor{black}{The STD matrix for the future day $d$ could be predicted if we have historical delivery order data of past consecutive $k$ days. First we construct a series of STD matrices $\bf{E}^{d-1}, \bf{E}^{d-2}, ..., \bf{E}^{d-k}$ for past consecutive $k$ days according to Definition~\ref{def: std matrix}. Then the STD matrix $\bf{E}^d$ of day $d$ could be predicted in an element-wise way. Specifically, each element $e^d_{i,j}$ could be predicted by:
\begin{equation}
    e^d_{i,j} = G(\{e^{d'}_{i,j}| d'=d-1, d-2, ..., d-k\}),
\end{equation}
where $\{e^{d'}_{i,j}| d'=d-1, d-2, ..., d-k\}$ is a time series drawn from historical STD matrices and $G$ is an aggregate function applied on the time series to get prediction. Here for efficiency of inference, we just take the average function as $G$. However, to get more accurate prediction, advanced spatial-temporal prediction methods~\cite{zhang2016dnn, pan2019urban} could be directly applied on these STD matrices.}

\subsubsection{Spatial-temporal score}
\textcolor{black}{
Practices in Didi Chuxing~\cite{wang2018deep, zhang2017taxi} suggest that `hitchhiking' could effectively optimize the logistics cost due to the reduction of travel length. Thus in DPDP, the delivery capacities are supposed to be consistent with delivery demands from both the spatial and temporal perspectives. In this way, vehicles could cover more delivery demands along planned route with less marginal logistics cost. To this end, Spatial-Temporal Score (ST Score) is proposed. For vehicle $k$ and its route $rp_k$, ST Score is utilized to measure the difference between delivery capacities and predicted delivery demands along the route $rp_k$. The following definitions are given to formulate the calculating procedure.
}
\textcolor{black}{
\begin{definition}[Spatial-temporal Coordinate] The traveling time condition is simplified by considering a constant average speed when moving from one node to another and the distance between any pair of nodes (factories or depots) is known. Based on that, we could estimate the time interval $TI_j$ that vehicle $k$ arrives at factory $F_i$ of route $rp_k$, given the departure time of vehicle $k$ and the loading and unloading time at factories. $(i,j)_{F_i,TI_j}$ denotes the spatial-temporal coordinate for vehicle $k$ arriving at $F_i$ within time interval $TI_j$.
\end{definition}
\begin{definition}[Spatial-temporal Capacity Vector] $\eta_i^k$ represents the residual capacity of vehicle $k$ arriving at factory $F_i$ of route $rp_k$, which is calculated by:
\begin{equation}
    \eta_i^k=Q+\sum_{h} q_+^h -\sum_{h'} q_-^{h'}
\end{equation}
where $q^{h}_+$ and $q^{h}_-$ respectively represent the unloading and loading amount of cargoes before factory $F_i$. In this way, a spatial-temporal capacity vector $\bm{\eta^k}=\{\eta_i^k|F_i \in rp_k\}$ could be constructed for route $rp_k$.
\end{definition}
\begin{definition}[Spatial-temporal Demand Vector] Given a predicted STD matrix $\bf{E}$ $=[e]_{n\times T} \in \mathbb{R}^{n\times T}$ and a series of spatial-temporal coordinates $\{(i,j)_{F_i, TI_j}\}$ for route $rp_k$, a predicted spatial-temporal demand vector $\bm{\tau^k} =\{\tau_i^k|(i,j)_{F_i, TI_j}\}$ for route $rp_k$ could be constructed via extracting corresponding element according to the spatial-temporal coordinates from $\bf{E}$, namely $\tau_i^k = e_{i,j}$.
\end{definition}
\begin{definition}[ST Score] $\xi_k \in \mathbb{R}$ denotes the ST Score of route $rp_k$ \textit{w.r.t.} predicted delivery demand, calculated by:  
\begin{equation}
    \xi_k = D_{JS}(\bm{\eta^k}|| \bm{\tau^k})
    \label{eq: JS}
\end{equation}
where $D_{JS}$ is the \textit{Jensen-Shannon} (JS) divergence function. 
\end{definition}
}
\begin{figure}[t]
	\centering
	\includegraphics[width=0.8\linewidth]{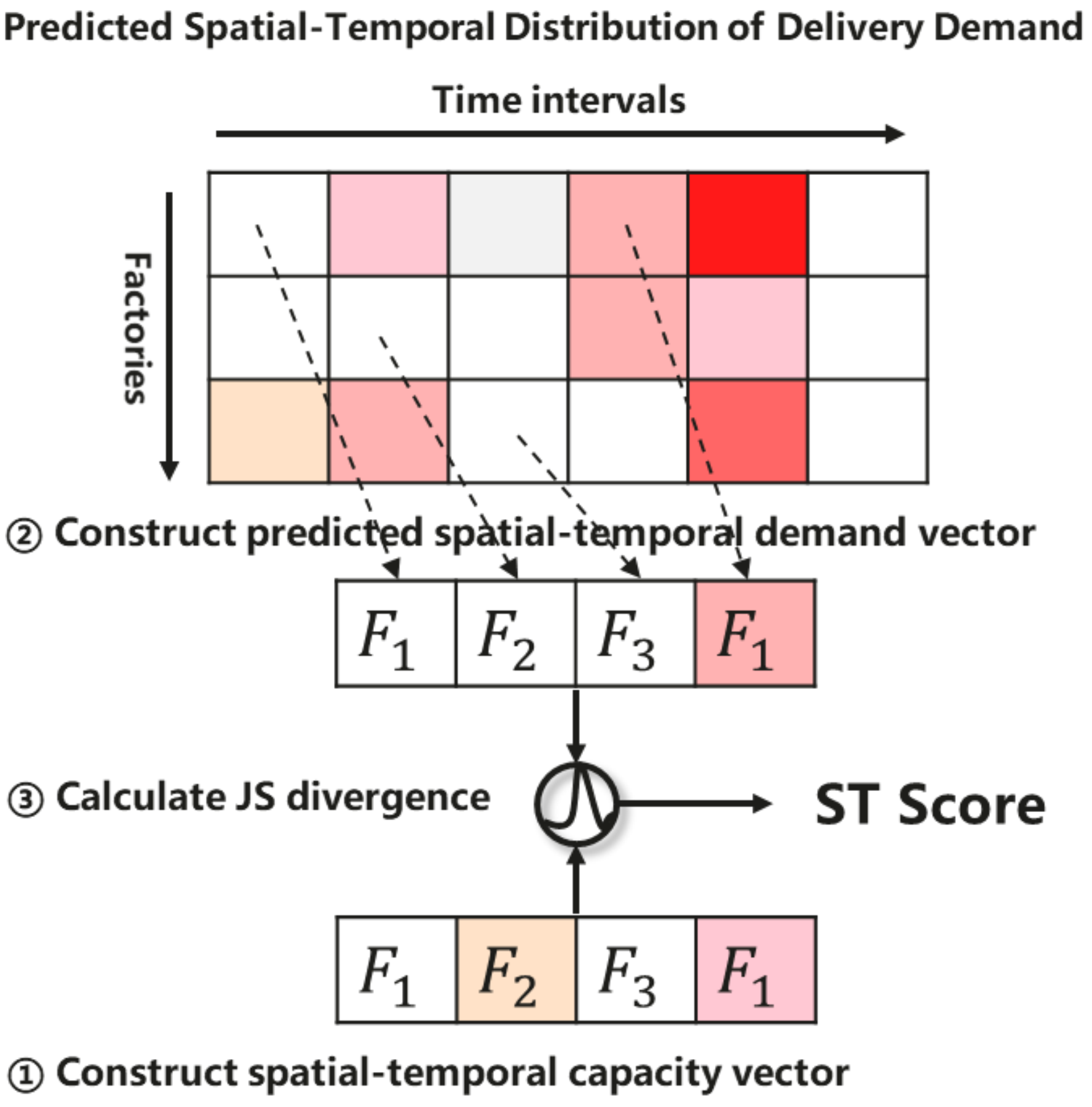}
	\caption {How to calculate the Spatial-Temporal Score (ST Score) for route (vehicle) with respect to predicted delivery demand. We first construct the spatial-temporal capacity vector according to vehicle's ongoing route. Similarly, the spatial-temporal demand vector could be constructed via selecting corresponding elements from predicted STD of delivery demand. Finally, we can obtain the ST Score by calculating the \textit{Jensen-Shannon} (JS) divergence between the two vectors.}
	\label{fig: st_score}

\end{figure}

\textcolor{black}{Various divergence metrics (such as KL and JS divergence) could be applied to measure the difference of two vectors in Eq.(\ref{eq: JS}). The reasons why the JS divergence is adopted here are 1) that the JS divergence is naturally symmetric in calculation and 2) that the experimental results show that JS divergence performs slightly better than symmetric KL divergence in our solution, given in Section~\ref{sec: supplementary materials}.}
Intuitively, the smaller the ST Score, the higher the probability of `hitchhiking' occurring. The whole calculating procedure is depicted in Fig.~\ref{fig: st_score}. 
\vspace{-2ex}
\subsection{Markov decision process}
\label{MDP}

As defined in Section~\ref{sec: problem}, a delivery order specifies two nodes, i.e., the pickup node and the delivery node. And a route speciﬁes the traverse sequence of the associated vehicle according to its accepted delivery orders. For a delivery order, dispatching which vehicle to serve the order is equivalent to selecting the route of which vehicle such that the two nodes of the order can be inserted. Hence, a \textit{route-centric Markov} Decision Process (MDP) formulation is given below.





\textbf{Objective}: As illustrated in Section~\ref{sec: problem}, the objective of our DRL solution to DPDP is to minimize the total transportation cost for serving all the delivery orders.

\textbf{State}: $S^i_t=(s^i_{t,1}, s^i_{t,2}, ..., s^i_{t,K})$, where $S^i_t$ is the joint state of the fleet with respect to order $o^{i}_{t}$, and $s^i_{t,k}$ is the individual state of the route $rp_k$ of vehicle $k$ with respect to $o^{i}_{t}$. The subscript $t$ here specifies that the order is generated within time interval $t$~\footnote{For conciseness of notation, we here denote time interval by $t$ instead of $TI$ used in Section~\ref{subsec: st_distribution}}. Suppose that vehicle $k$ is feasible to take order $o_t^i$. Then we could have at least one temporary route $rp_{t,k}^i$ resulting from inserting the two nodes of $o^i_t$ into $rp_k$. For the individual state $s^i_{t,k}$, it includes three major entries. The first one is the length of route $rp_k$, which is denoted by $d_{t, k}$. The second one is the length of $rp_{t,k}^i$, denoted by $d^i_{t,k}$. The last one is the ST Score $\xi^i_{t,k}$ of $rp_{t,k}^i$. Besides, a used flag $f_{t,k}$ and the current time interval $t$ are also added into $s^i_{t,k}$. The former represents whether the vehicle $k$ is used before. If it has served any delivery order, $f_{t,k}=1$, otherwise $f_{t,k} = 0$. In a word, $s^i_{t,k}$ is formulated as $s^i_{t,k}=(d_{t, k}, d^i_{t,k}, \xi^i_{t,k}, f_{t,k}, t)$ and $S_t^i \in \mathbb{R}^{K\times5}$. The detailed procedure of obtaining state is given in Algorithm~\ref{alg: route planner} and Algorithm~\ref{alg: ddgn}.

\textbf{Action}: $a^{i}_{t}$. For a given order $o_t^i$, $a^{i}_{t}$ represents the action that a vehicle is selected to serve the order. Thus the action space $\mathcal{A}$ is discrete and with the dimension of $K$.


\textbf{Reward}: $R^i_t\in \mathbb{R}$. Once vehicle $k$ is chosen to serve order $o^i_t$, it gains an instant reward $r^i_t$, which is calculated by:
\begin{equation}
\label{eq: instant reward}
r^i_t = - \alpha \times (\mu \times f_{t,k} + \delta \times \triangle{d}^i_{t,k})
\end{equation}
where $\alpha$ is a reward factor, $\mu$ is the fixed cost of using a new vehicle, $\delta$ is the operation cost of a vehicle per kilometer and $\triangle{d}^i_{t,k}=d^i_{t,k}-d_{t,k}$ represents the incremental distance caused by adding nodes of order $o^i_t$ into vehicle $k$'s route. Besides, a long-term reward $\bar{r}$ is considered to prevent the learned policy from being too myopic, which is calculated by:
\begin{equation}
\label{eq: long term reward}
\bar{r}=\frac{\sum_{t=1}^{T}\sum_{i=1}^{M_t} r^i_t}{\sum^T_{t=1} M_t}
\end{equation}
where $T$ is the total number of time intervals in an episode (i.e., a $24$-h day) and $M_t$ is the number of orders served in time interval $t$. Note that the long-term reward is calculated at the end of the episode. So the reward $R^i_t$ is calculated by:
\begin{equation}
\label{eq: final reward}
R^i_t = r^i_t + \bar{r}
\end{equation}

\textbf{State transition}: $p(S^{i^\prime}_{t^\prime} | S^i_t, a^{i}_{t})$. Once the order $o_t^i$ is assigned to one vehicle, there is a state transition from the current state $S^i_t$ to the next state $S^{i^\prime}_{t^\prime}$. In the next state $S^{i^\prime}_{t^\prime}$, $t^\prime$ might still equal to $t$ since the next delivery order might still fall into time interval $t$. It should be noted that the state transition is stochastic even though the influence of taking action $a^{i}_{t}$ is deterministic. Because the next state $S^{i^\prime}_{t^\prime}$ is related to the next order, which is unknown currently.




\vspace{-2ex}
\subsection{Relational learning framework}


Based on above MDP formulation, we propose a relational reinforcement learning framework aided by spatial-temporal prediction. The framework is described below.

\subsubsection{ST-DDGN}

\label{subsec: double dgn}



\begin{figure}[t]
	\centering
	\includegraphics[width=0.95\linewidth]{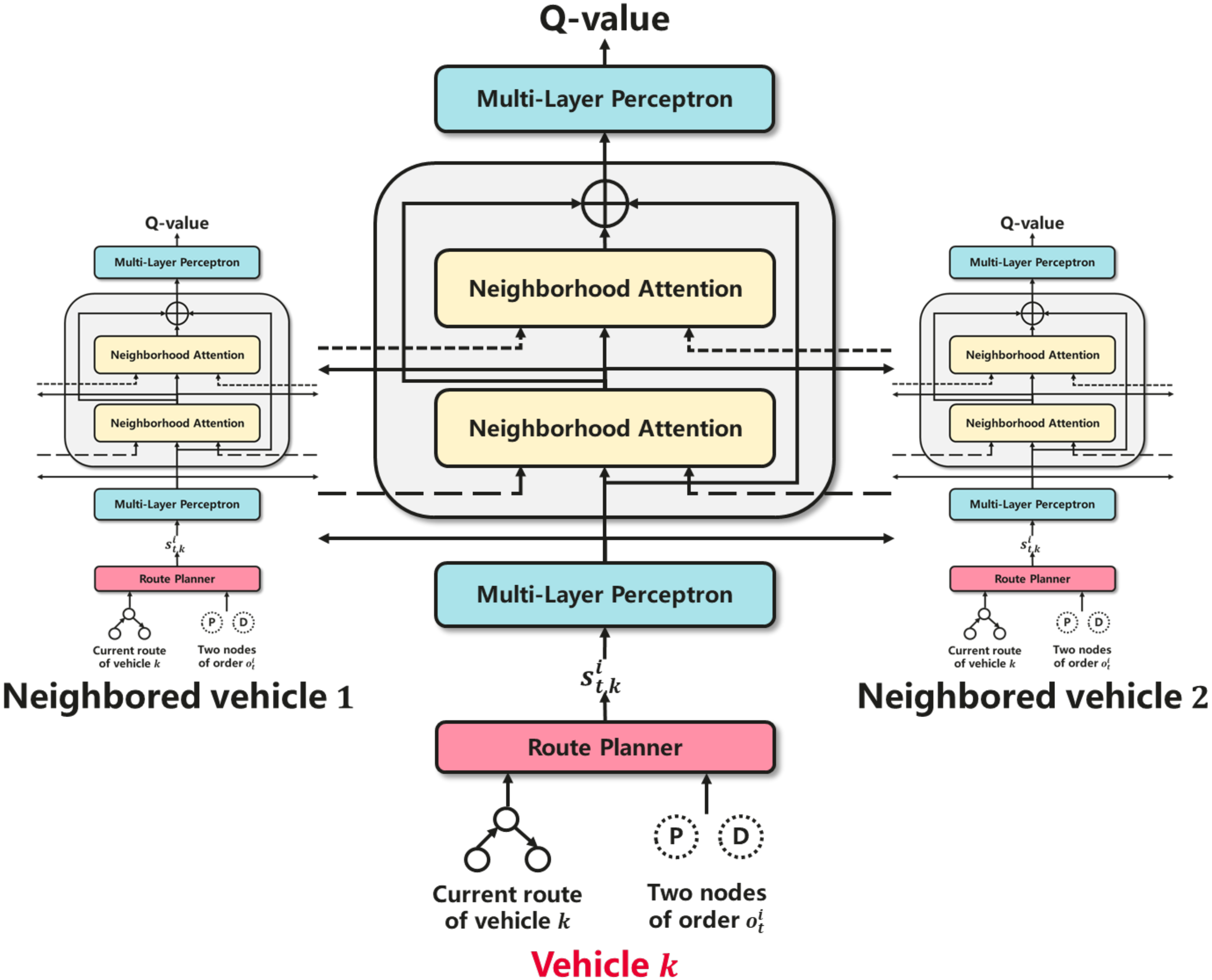}
	\caption{Architecture of ST-DDGN. The network consists of three main components: route planner, multi-layer perceptron (MLP), neighborhood attention. Each vehicle owns its network but shares the same weights. Multiple such networks comprise the proposed ST-DDGN.}
	\label{fig:DGN}
\end{figure}

As described in Section \ref{subsec: st_distribution}, we utilize the ST Score to guide the policy learning, making it possible to dispatch the fleet to the `hot spot' of demand. Besides, graph convolution is often adopted to learn the relation between entities in the environment~\cite{zambaldi2018relational, jiang2018graph}. Motivated by the two perspectives, we propose Spatial-Temporal Aided Double Deep Graph Network (ST-DDGN). The dynamic dispatching environment is regarded as a graph and each vehicle in the environment is abstracted as a node in the graph. As shown in Fig.~\ref{fig:DGN}, ST-DDGN consists of three main components: route planner, multi-layer perceptron (MLP), and neighborhood attention. Roughly speaking, for a given delivery order $o_t^i$, ST-DDGN first calculates a Q-value for each vehicle $k \in K$ taking the order, which evaluates how good assigning order $o_t^i$ to vehicle $k$ is. And then the order is finally assigned to the vehicle with the highest Q-value. More specifically, the route planner is responsible for constructing the state $s_{t,k}^{i}$ for vehicle $k$ and order $o_{t}^{i}$. Then the state $s_{t,k}^{i}$ is sent to the initial MLP to get an initial representation. In order to obtain the high-level representation of interaction between vehicle $k$ and other vehicles, a followed neighborhood attention module integrates the initial representations within a predefined range, namely neighborhood. In other words, the neighborhood attention module functions as a communicator among the fleet of vehicles. Note that the neighborhood attention modules could be stacked. 
At last the initial and high-level representations are concatenated and passed on to the final MLP which gives out the Q-value. Note that the policy network in the proposed framework is not limited on DDQN. Other advanced policy gradient methods, such as A3C~\cite{mnih2016a3c} and DDPG~\cite{2015Continuous}, could also be incorporated into our framework. In the following, the characteristics of DDGN are introduced at length.

\textbf{Constraint embedding}. To accommodate prior constraints, contextual DQN~\cite{lin2018efficient} shrinks the output action space with contextual masks based on prior rules. It wastes computational resources if ST-DDGN adopts such a way to filter feasible vehicles. For a given order, not every vehicle is feasible to take the order since some of them violate one or more constraints. Hence it is not necessary to make inference for each vehicle. Besides, due to the architecture of ST-DDGN which could be seen as a factorization of centralized decision maker, the inference time increases as the total number of vehicles involved becomes large. Thus we could exclude infeasible vehicles at early stage of inference. The route planner module takes this job. For a given delivery order $o_t^i$ and a vehicle $k$, the route planner is responsible for 1) checking whether the vehicle $k$ is feasible to take order $o_t^i$; 2) planning a shortest route for vehicle $k$ taking order $o_t^i$ if it is feasible; 3) calculating entries of state $s_{t,k}^{i}$ defined in Section \ref{MDP} if it is feasible. For more details about the route planner, please see the pseudo code in Algorithm~\ref{alg: route planner}.

The feasible and infeasible vehicles can be distinguished using the route planner. For those infeasible vehicles, they are not involved in the following inference, where the weight connections related to the state are cancelled. Besides, the cancellation would lead to no error generated in back propagation, which means that the infeasible vehicle makes no contribution to the final learned policy. Note that we directly set the Q-value of infeasible vehicles to extremely small negative, ensuring that these vehicles have no chance to be chosen.

\begin{figure}[t]
	\centering
	\includegraphics[width=0.84\linewidth]{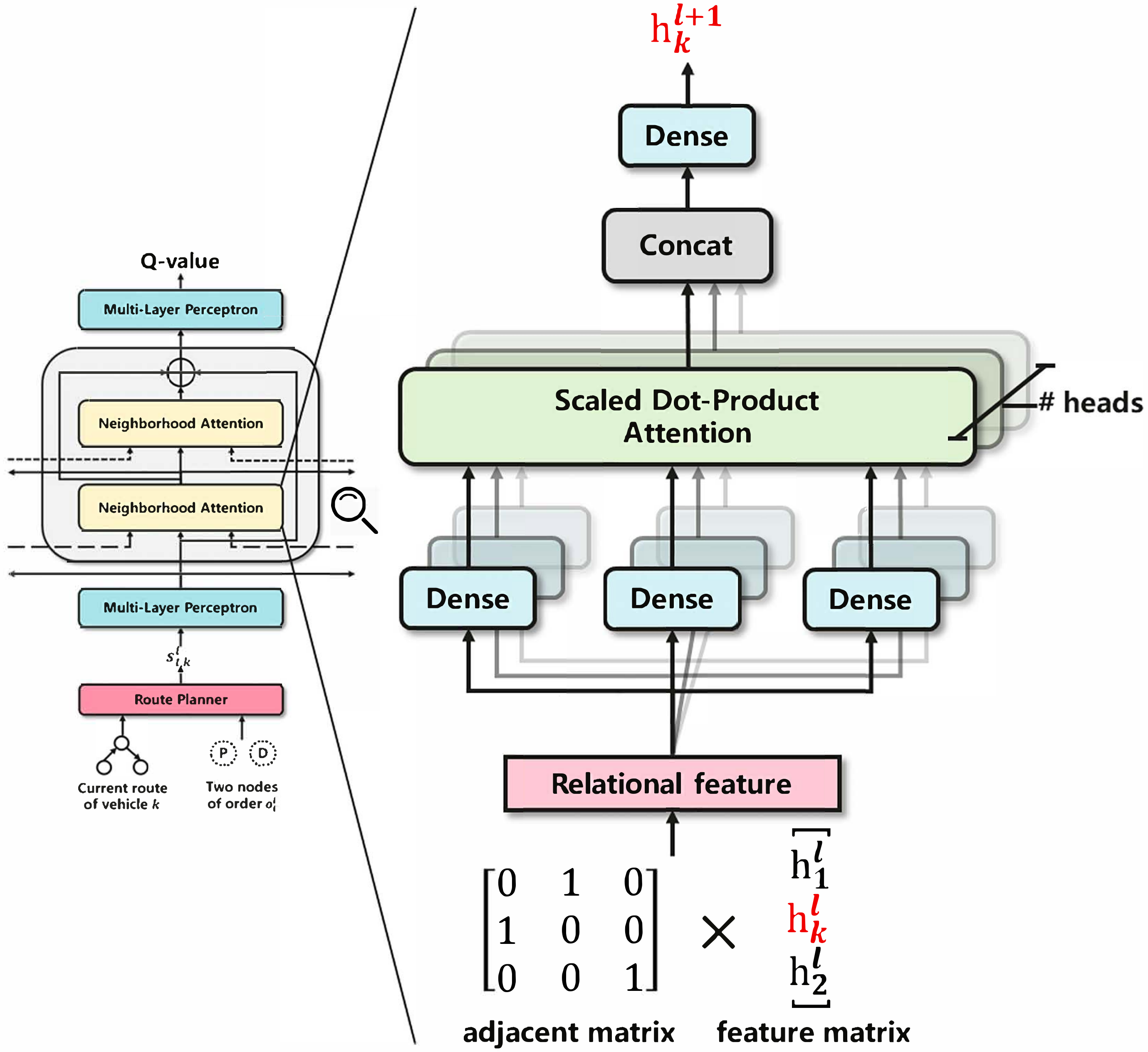}
	\caption{Neighborhood attention module. For a given vehicle $k$ at level $l$, it receives lower level representation (i.e., feature matrix) from its real-time neighbored vehicles. A relational feature is then a product of the feature matrix and adjacent matrix. The relational feature is then sent to multi-head scaled dot-product attention, followed by a Dense layer. In this way, a higher level representation (i.e., $h_k^{l+1}$) could be obtained for vehicle $k$ and its neighbors.}
	\label{fig: attention}
\end{figure}

\textbf{Neighborhood attention}. As shown in Fig.~\ref{fig: attention}, each attention block corresponds to one vehicle (e.g., vehicle $k$). The block at level $l+1$~\footnote{Here the level is specially defined for neighborhood attention and starts from $0$. But level $0$ refers to initial representation obtained from the first MLP.} receives lower level representations (i.e., $h_1^{l},...,h_k^{l},...,h_K^{l}$) from both vehicle $k$ and other vehicles. These lower level representations together comprise a feature matrix.
In additions, an adjacent matrix with respect to vehicle $k$ is also taken as input into the block. More specifically, each row of the adjacent matrix is an one-hot vector. The length of the vector is $K$ and the position of $1$ in the vector refers to a neighbored vehicle of vehicle $k$. Here we adopt the \textit{Euclidean} distance to measure the spatial proximity between vehicles and select $N\!E$ nearest vehicles with respect to vehicle $k$ as neighbors. Thus the adjacent matrix is of $\{0,1\}^{N \! E \times K}$. A relational feature is the product of the feature matrix and adjacent matrix.
Then the relational feature is sent to multi-head scaled dot-product attention~\cite{vaswani2017attention}, followed by a Dense~\cite{huang2017densely} layer. In this way, a higher level representation (i.e., $h_k^{l+1}$) could be obtained for vehicle $k$ and its neighbors. The corresponding hyperparameters of the neighborhood attention block are given in Section~\ref{sec: supplementary materials}.
	
The neighborhood block could be stacked to multiple levels. The connection weights are learned in an end-to-end way and shared by all attention blocks at the same level, equal to a spatial convolution over the graph. With the stacking of multiple attention blocks, one attention block at higher level can integrate more information through the graph convolution with larger receptive field. We stack two levels of blocks in this work owing to the trade-off between accuracy and efficiency.

\subsubsection{Simulator}
\label{sec: simulator}

Simulator is of crucial importance in our dispatching system, responsible for the simulation of interaction between agent and environment. The historical data is fed into the simulator to simulate how our dispatching system runs in real world. Algorithm \ref{alg: simulator} presents the details how the simulator runs in one \textit{episode} (i.e., one 24-h day).

At the initial phase, the delivery orders of one episode are sampled uniformly from a historical data pool. Then these orders are sorted in the ascending order of creation time $t_c^i$ and partitioned into $T$ time intervals of the same duration. It should be noted that the time interval is just used to assign the time feature to each delivery order instead of buffering orders. Vehicles starting at various depots are initialized following the actual spatial distribution. Within each time interval $t$, the information, such as the route, current location and residual capacity, of all vehicles is updated first. For each unassigned order within the time interval $t$, the most `appropriate' vehicle is selected by the learned policy to serve the order. After that, the information of the selected vehicle is further updated due to serving the order, including the route and route length.



\begin{table*}[t]
	\caption{Performance comparison between DRL methods and optimal solution}
	\label{tab:optimal}
	\centering
	\small{
	\begin{tabular}{c|ccc|ccc|ccc|ccc}
		\hline
		\multirow{2}{*}{Algo.} & \multicolumn{3}{c|}{6 orders} & \multicolumn{3}{c|}{7 orders} & \multicolumn{3}{c|}{8 orders} & \multicolumn{3}{c}{10 orders} \\ \cline{2-13} 
		& NUV   & TC      & wall time  & NUV   & TC      & wall time  & NUV   & TC      & wall time  & NUV   & TC      & wall time   \\ \hline
		DQN                   & 4     & 1415.02   & 0.14s     & 5     & 1763.02   & 0.2s      & 5     & 1891.92   & 0.22s     & 5     & 1953.6   & 0.25s      \\
		AC                  & 4     & 1415.02   & 0.14s     & 4     & 1468.18   & 0.2s      & 5     & 1891.92   & 0.22s     & 5     & 1953.6   & 0.25s      \\
		DGN                   & 3     & 1188.2    & 0.23s     & 4     & 1468.18   & 0.27s     & 4     & 1489.84   & 0.3s      & 5     & 1953.6   & 0.34s      \\
		ST-DDGN                  & 3     & 1188.2    & 0.23s     & 4     & 1468.18   & 0.27s     & 4     & 1489.84   & 0.3s      & 5     & 1953.6   & 0.34s      \\ \hline
		MIP                   & 3     & 1158.2    & 300.36s   & 3     & 1206.97   & 2818.14s  & $-$   & $-$      & $-$       & $-$   & $-$      & $-$        \\ \hline
	\end{tabular}
	}
\end{table*}

\subsection{Discussions}

\textcolor{black}{It should be noted that the processing order of delivery order does matter in DPDP. Previous work~\cite{pureza2008waiting} suggests that for DPDP, waiting and buffering strategy is effective in the reduction of logistics cost. However, we still adopt the immediate service strategy in this work due to short response time ($60$s) of business requirement. Besides we have tried a fixed time-interval buffering strategy in an early solution. The preliminary results suggest that the logistics costs are not reduced obviously compared with immediate service strategy but with much longer response time ($154.47$s on average per order). The variable time-interval buffering strategy with short response time requirement is worthy of being investigated further, which is left to our future work.}

\textcolor{black}{Besides, there might be a few unusual scenarios where the orders increase significantly. These special situations make rewarding the decision-making for minimizing the ST Score inappropriate because the STD of orders to be served differs greatly from that of historical orders. Hence, we introduce the ST Score into the state instead of the reward function. The dispatching system could thus take into account not only ST Score but also other factors (such as the incremental cost of dispatching an order), so as to make wiser and more robust decisions when encountering the unusual situations.} 

\section{Experiments}
\label{sec: experiment}
\subsection{Experimental methodology \label{sec: metric}}

We compare our ST-DDGN with a state-of-the-art heuristic algorithm, a mixed integer programming method (exact algorithm) and a variety of vanilla DRL methods.

\textcolor{black}{\textbf{Baselines:} Minic \textit{et al}.~\cite{mitrovic2004waiting} propose a common strategy to solve DPDP, i.e., inserting the nodes of incoming order into the route of a specific vehicle. 
We adapt previous works \cite{mitrovic2004waiting, Grandinetti2014the} to three baselines that dispatch each order to the most `appropriate' vehicle based on various greedy rules:
\begin{itemize}
    \item Baseline 1 adopts the strategy in~\cite{mitrovic2004waiting}, dispatching a new order to the vehicle that has the shortest incremental route length after accepting the order. This baseline aims at minimizing the operation cost.
    \item Baseline 2 is also inspired by~\cite{mitrovic2004waiting}, dispatching the order to the vehicle with the shortest total route length after accepting the order. Similar to baseline 1, this strategy would also like to optimize the operation cost but from another perspective. 
    \item Baseline 3 follows a strategy adapted from~\cite{Grandinetti2014the}, which aims at reducing the fixed cost via reducing the number of used vehicles. Specifically, this algorithm dispatches the order to the vehicle that has the largest number of accepted orders.
\end{itemize}
Note that the baseline 1 has been deployed for months in the UAT environment of our company's logistics system. By comparing with above baselines, it could be estimated how much cost will be saved if ST-DDGN is fully deployed in the production environment.}

\textbf{Mixed integer programming:} If all information about the delivery orders is known a priori, DPDP is reduced to a pickup and delivery problem (PDP), a static version of DPDP. We give a three-index mathematical formulation in Section~\ref{sec: supplementary materials} for PDP by assuming all information are known beforehand. Then by using a solver to solve the corresponding mixed integer programming (MIP) model, we can get the optimal solution. By comparing the result obtained from our method with that from solving the MIP model, we could roughly know how far away our solution is to the global optimum. It should be noted that the MIP method can only be used to solve small-scale PDP due to the NP-Hardness of the problem.

\textbf{Vanilla DRL methods:} DQN~\cite{mnih2015human} and Actor-Critic (AC)~\cite{ac2012} are classic model-free DRL frameworks that has achieved impressive performance in many scenarios. In additions, DGN \cite{jiang2018graph} is a graph-based relational reinforcement learning method which is good at capturing the correlation between individuals. We apply these vanilla DRL methods, using the MDP formulation defined in Section~\ref{sec: solution} except the ST Score, to solve DPDP. We compare our ST-DDGN with them to verify its advantages over these vanilla DRL methods.

On the basis that all delivery orders can be served, we compare the above algorithms using the following metrics:


\textbf{Number of the Used Vehicles (NUV)}. NUV is the total number of used vehicles to serve all the delivery orders in one episode. The less NUV, the better the dispatching algorithm is. Note that NUV cannot exceed the predefined maximum number of vehicles, i.e., $K$.

\textbf{Total Cost (TC)}. The total cost $TC$ is calculated by $TC= \mu \times \textrm{NUV}$ $+$  $\delta \times TTL$ where $TTL$ is the total travel length of all used vehicles in one episode.

\subsection{Setting up}
\label{subsec: setting}
Since the actual traveling time of vehicles under dynamic environment cannot be gotten offline, \textcolor{black}{a simulation environment is built to simulate the dispatching process within Huawei's manufacturing campus located in Pearl River Delta region, China. This campus consists of $27$ factories where hundreds of delivery orders are generated every day.} The distance between any pair of factories in the environment is identical to that in real world. Traveling time condition is simplified by considering a constant vehicle average speed when moving from one factory to another. A fleet of homogeneous vehicles is initialized at various depots following the actual geographical distribution of vehicles. 

All experiments are conducted on data extracted from practical scenarios. \textcolor{black}{We collect historical data (delivery orders from July to October 2019) as the integrate dataset which contains around 80,000 delivery orders. It should be noted that orders of the last 20 days of October 2019 are regarded as testing data and the rest of orders are the training data. We would like to verify the generalization of ST-DDGN by splitting up the integrate dataset since the spatial-temporal distribution of testing data is different from that of training data.} Furthermore, various scales of training and testing instances are constructed by uniformly sampling from the testing and training data respectively. Thus, the spatial-temporal distribution of orders in these instances are independent from each other while identical to that of real historical order data. 

In order to reduce the randomness of training and to avoid the bias in sampling, the policy learning of DRL methods are conducted five times on each testing instance. The MIP and baselines are only performed on each testing instance once because there is no randomness in them. Besides, the recommended parameter/hyperparameter setting for various DRL methods is given in Section~\ref{sec: supplementary materials}.

\vspace{-1ex}
\subsection{Results and analysis}

\subsubsection{Performance comparison with optimal solution}

The MIP is used to get an optimal solution of DPDP by assuming an ideal case where all orders information within a day can be known beforehand. Due to extremely high computation complexity of MIP, only four tiny-scale testing instances are studied, where $5$ vehicles serves $6,7,8$ and $10$ orders respectively. The comparison between DRL methods and optimal solution is presented in Table~\ref{tab:optimal}.

As seen in Table~\ref{tab:optimal}, the ST-DDGN and DGN achieve the best solution quality among four DRL methods in all testing instances. When serving $6$ orders, the two graph-based DRL methods converge to a solution close to the optimum. In terms of the wall time, even in the smallest scale of instance ($6$ orders), MIP spends $300.36$s on finding the optimal solution. When serving more than $8$ orders, the MIP method is intractable (the result cannot be obtained within two hours even using the world fastest commercial solver Gurobi while the inference time of all DRL methods is less than $1$s.


\subsubsection{Performance comparison on large-scale instances}

\begin{figure}[t]
	
	\centering
	\subfigure[NUV]{
		\includegraphics[width=0.46\linewidth]{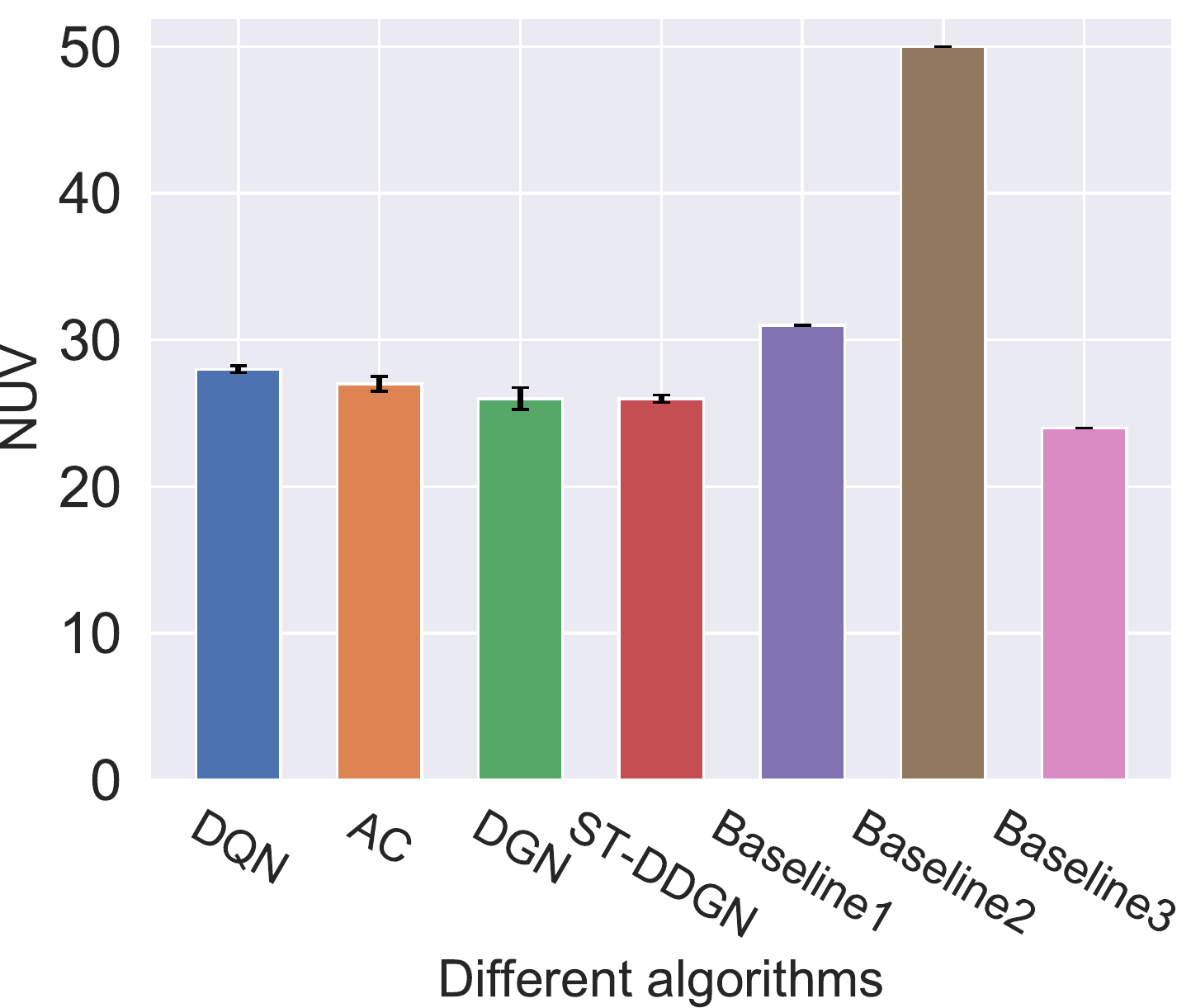}
		\label{fig: nuv-bar}
	}
	\subfigure[TC]{
		\includegraphics[width=0.472\linewidth]{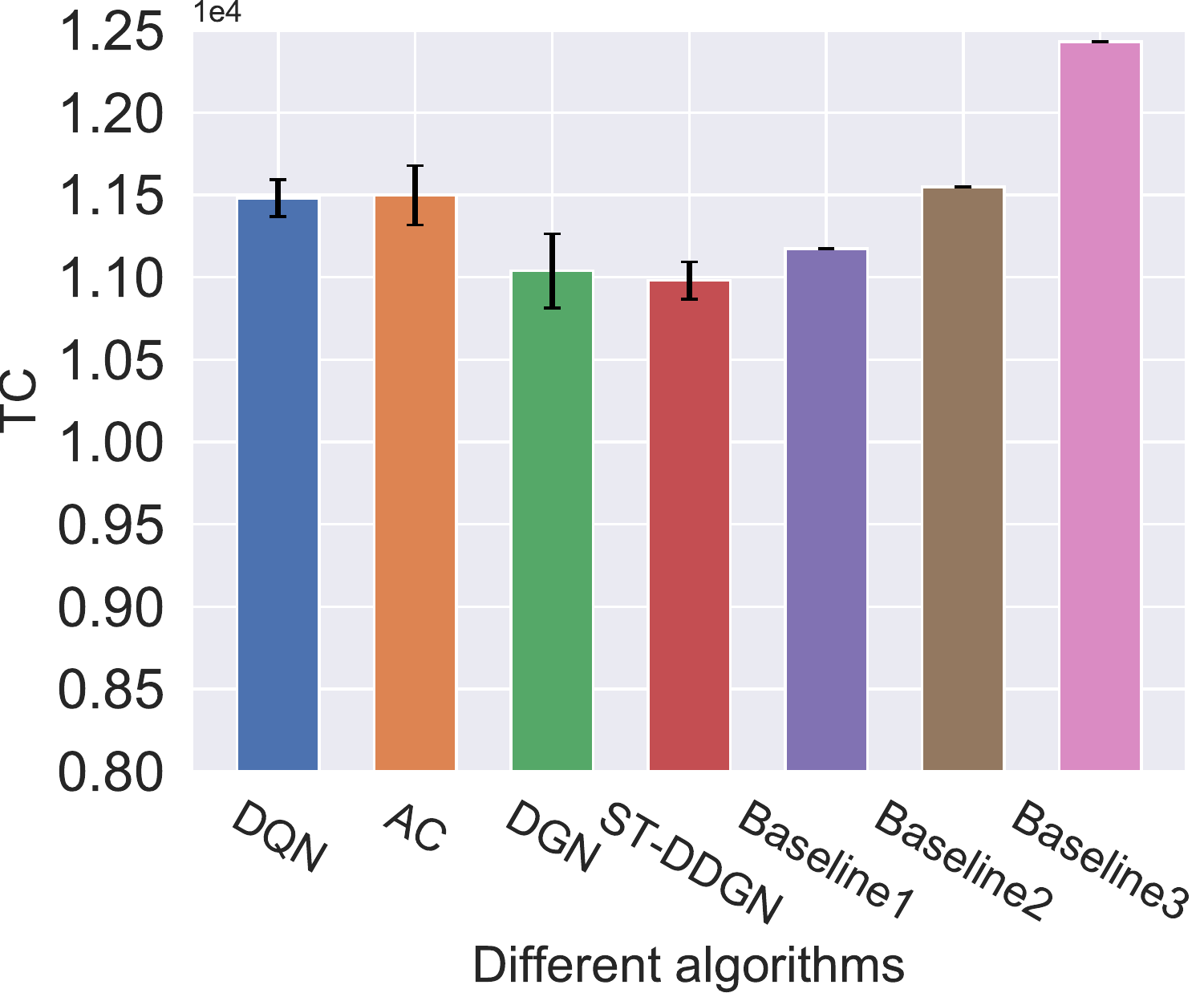}
		\label{fig: moc150-bar}
	}
	\caption{Comparing results of different algorithms on large-scale instances}
	\label{fig:Comparison}
\end{figure}

This experiment is conducted on multiple large-scale instances where $50$ vehicles are dispatched to serve $150$ delivery orders. Fig.~\ref{fig:Comparison} records the results on NUV and TC. In terms of NUV, baseline 3 performs best on all instances. ST-DDGN stands behind baseline 3, using $26.4$ vehicles on average.
\textcolor{black}{Besides, other DRL-based methods also perform much better than baseline 1 and baseline 2. With regards to TC, baseline 2 costs much less than baseline 3 even though it runs out of all vehicles. The result indicates that pursuing the least used vehicles blindly (i.e., baseline 3) might bring higher operation cost. Among the three baselines, baseline 1 achieves the least TC ($11167.5$). Graph-based DRL methods (DGN and ST-DDGN) outperform all three baselines. Especially, our ST-DDGN shows a certain advantage ($11080.5 \pm 112.5$) over the best baseline.}



%

\begin{figure*}[htbp]
	\centering
	\subfigure[NUV]{
		\includegraphics[width=1.0\linewidth]{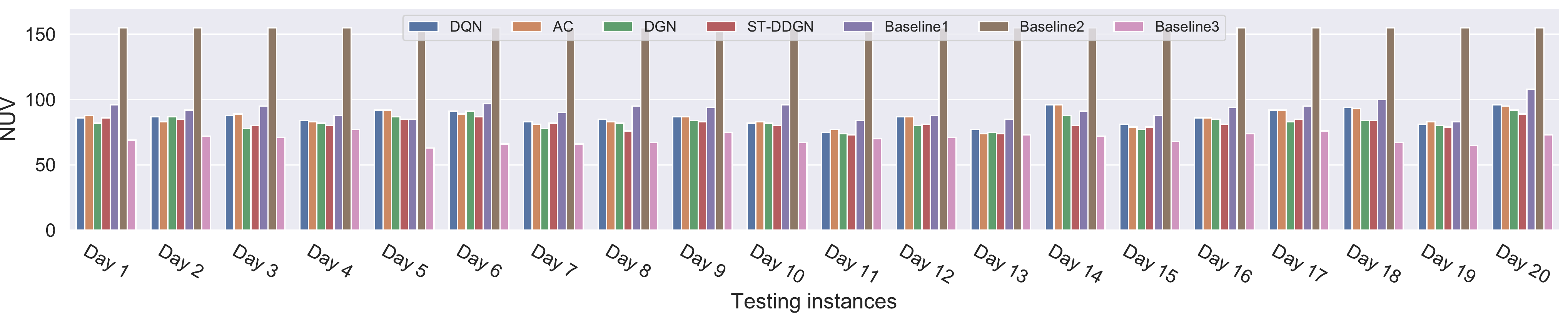}
		\label{fig:real-nuv}
	}
	\subfigure[TC]{
		\includegraphics[width=1.0\linewidth]{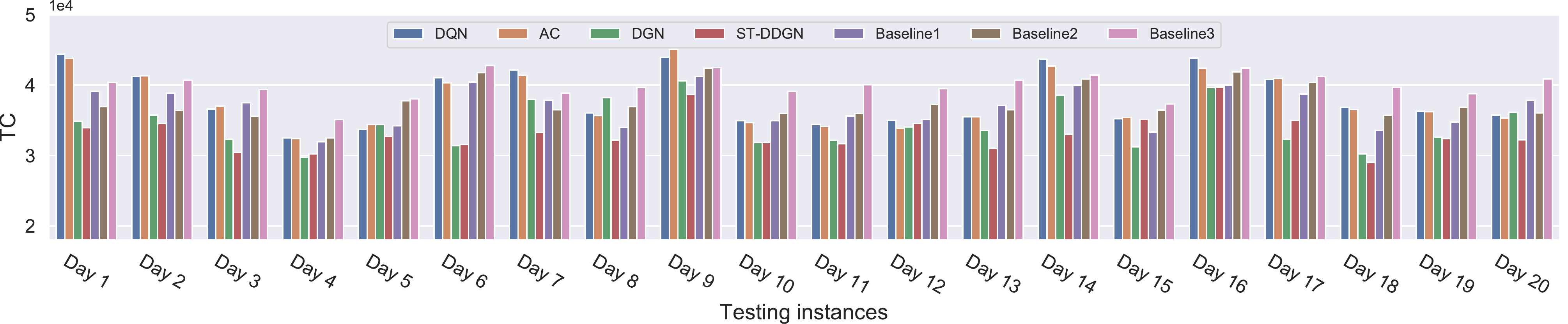}
		\label{fig:real-moc}
	}
	\caption{Comparing results of different algorithms on industry-scale instances}
	\label{fig: real}
\end{figure*}

\begin{figure}[t]
	
	\centering
	\subfigure[NUV curves]{
		\includegraphics[width=0.46\linewidth]{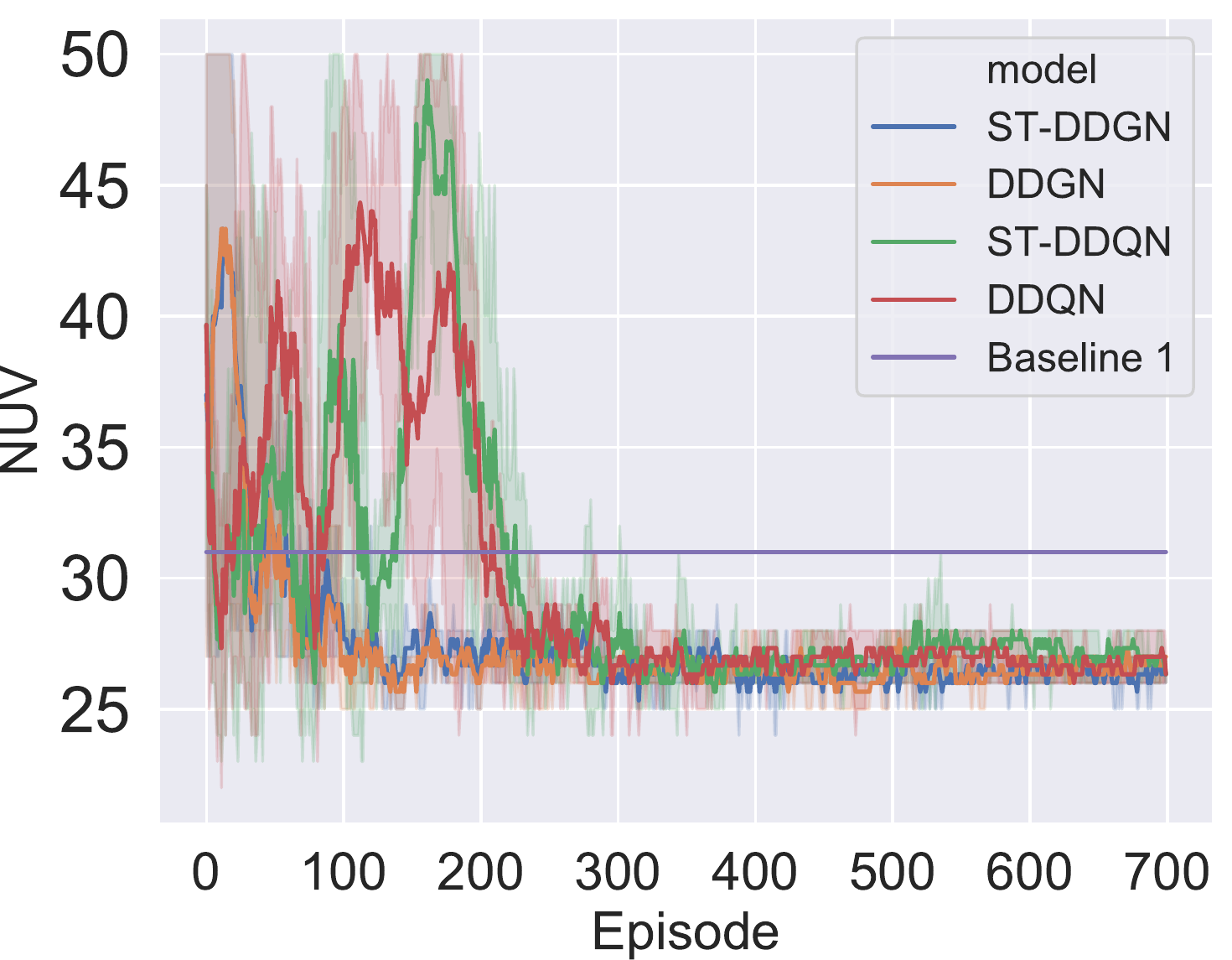}
		\label{fig:nuv-150}
	}
	\subfigure[TC curves]{
		\includegraphics[width=0.465\linewidth]{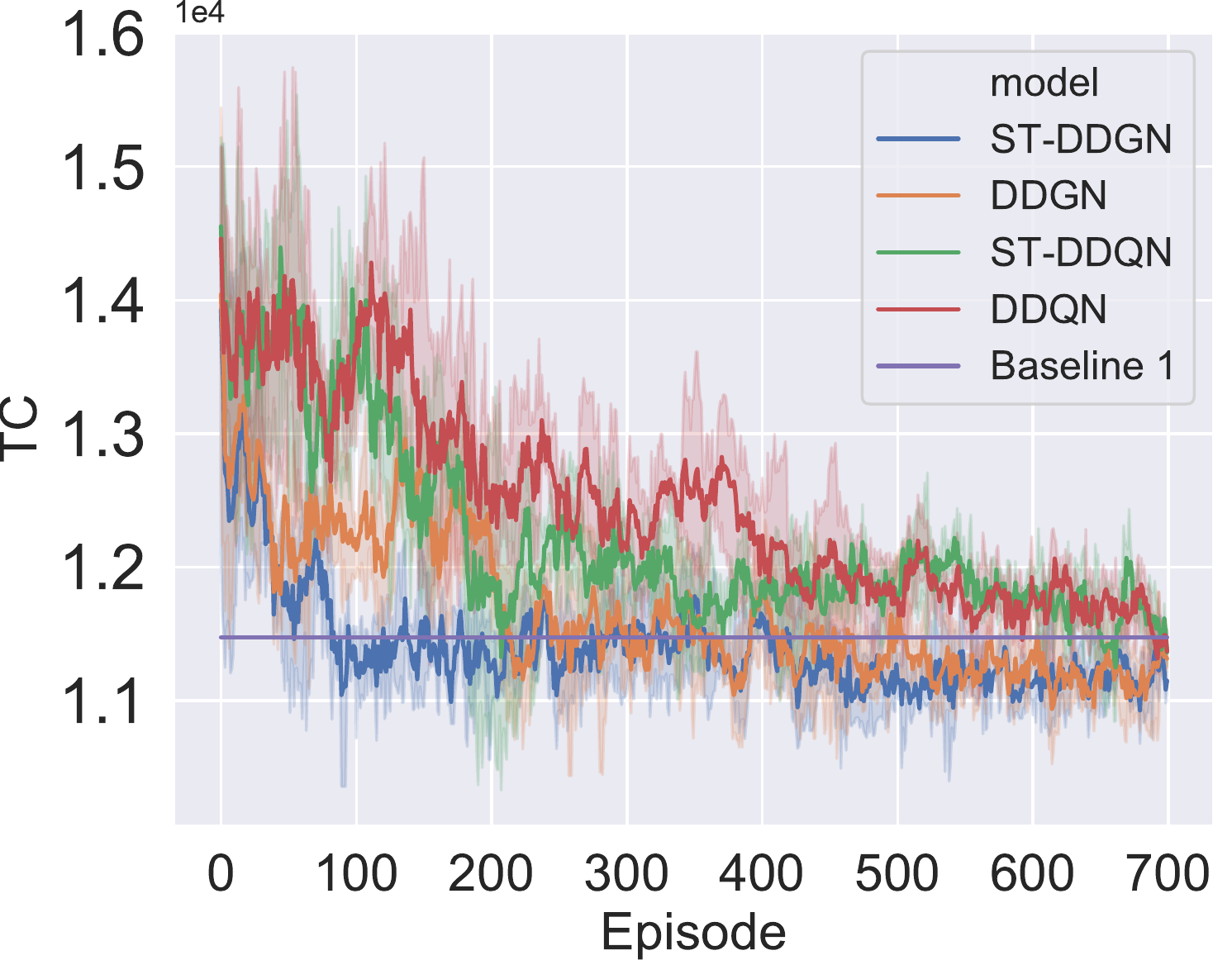}
		\label{fig:moc-150}
	}
	\caption{Convergence of different algorithms on large-scale instance}
	\label{fig:Sequential}
\end{figure}

\subsubsection{Performance comparison on industry-scale instances}

To verify the effectiveness of our ST-DDGN in real scenario, we further compare the performance between ST-DDGN, vanilla DRL methods and baselines on a group of industry-scale instances. Each instance involves $600+$ orders and $150+$ vehicles. Here, the `industry-scale' refers to that these instances are directly from the real daily transportation data (including delivery orders and cost setting) instead of the sampled instances used in previous experiments. Fig.~\ref{fig: real} presents the results, where the horizontal axis gives the index of testing instance (i.e., Day 1, Day 2, ... ) and vertical axis records NUV and TC respectively. \textcolor{black}{It can be observed from Fig.~\ref{fig:real-nuv} that baseline 2 uses all the vehicles and baseline 3 needs the least vehicles, which corresponds with the result of previous experiment. Correspondingly, baseline 1 performs the best among the three baselines. In comparison, DRL methods use less vehicles than baseline 1.} Specifically, for serving all delivery orders, baseline 1 uses $91.8$ vehicles on average while DRL methods only use $84.125$ vehicles on average, which justifies the effectiveness of our MDP formulation in reducing the number of used vehicles. While on TC, ST-DDGN surpasses all other algorithms on most of instances, where ST-DDGN costs $33162.7$ and baseline 1 costs $36812.8$ on average. Even in the instances where other algorithms perform better (such as Day 10 and Day 16), ST-DDGN could still achieve similar performance. \textcolor{black}{Thus, it is reasonable to consider that ST-DDGN can learn a policy with a trade-off between reducing the fixed cost and lowering the operation cost, which reduces the total cost effectively.} 




\subsubsection{Ablation studies}
\label{sec: ablation studies}

In this subsection, we would like to further justify the effectiveness of ST-DDGN via ablation studies. Compared to vanilla DRL methods, we introduce 1) the ST Score to balance `supply and demand' and 2) the graph convolution (i.e., neighborhood attention module) to grasp the interactions among vehicles. Thus, an ablation study concerning the two components above is performed. Specifically, four various versions of the proposed model are involved, presented in Table~\ref{tab: ablation studies}.
\begin{table}[ht]
	\caption{Model studied in ablation studies}
	\label{tab: ablation studies}
	\centering
	\small{
	\begin{tabular}{ccc}
		\hline
		Model & ST Score & Graph \\
		\hline
		DDQN  & $\times$ & $\times$ \\
		ST-DDQN & $\surd$ & $\times$ \\
		DDGN &$\times$&$ \surd$ \\
		ST-DDGN &$\surd$ &$\surd$ \\
		\hline
	\end{tabular}
	}
\end{table}

\begin{figure*}[ht]
    \centering
	\subfigure[DQN]{
		\includegraphics[width=0.42\linewidth]{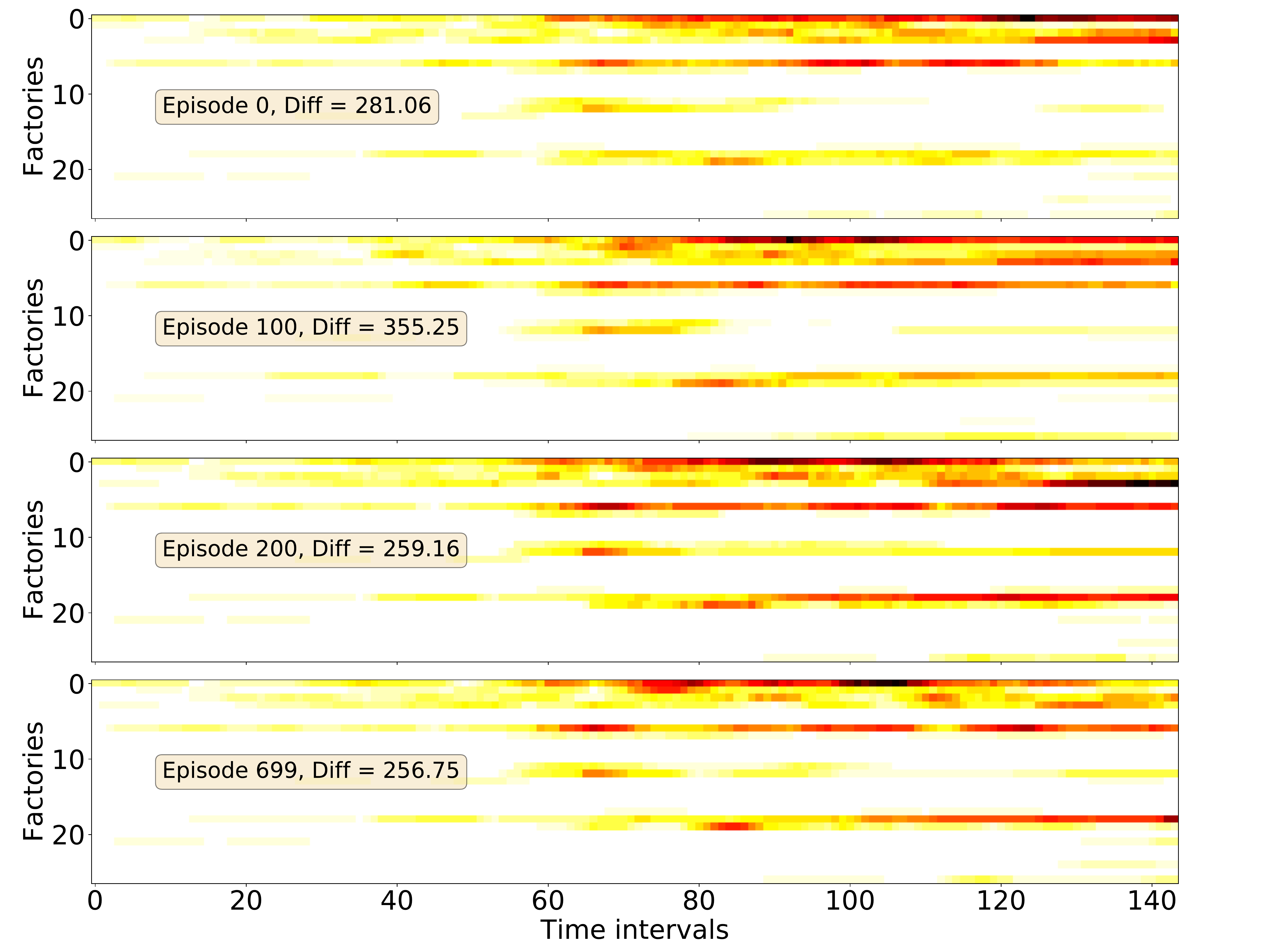}
		\label{fig: DQN iteration}
	}
	\subfigure[AC]{
		\includegraphics[width=0.42\linewidth]{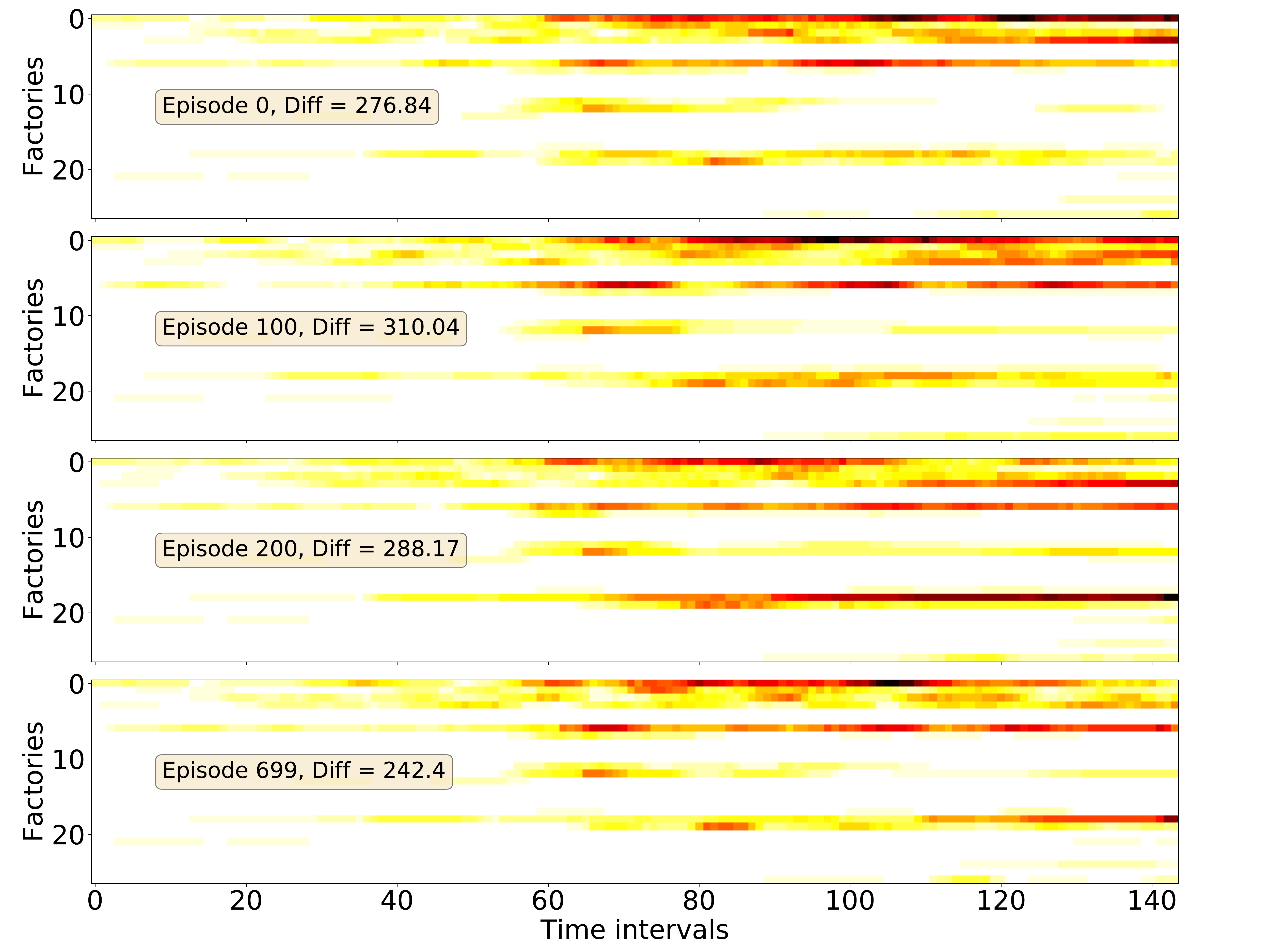}
		\label{fig: DDQN iteration}
	}
	\subfigure[DGN]{
		\includegraphics[width=0.42\linewidth]{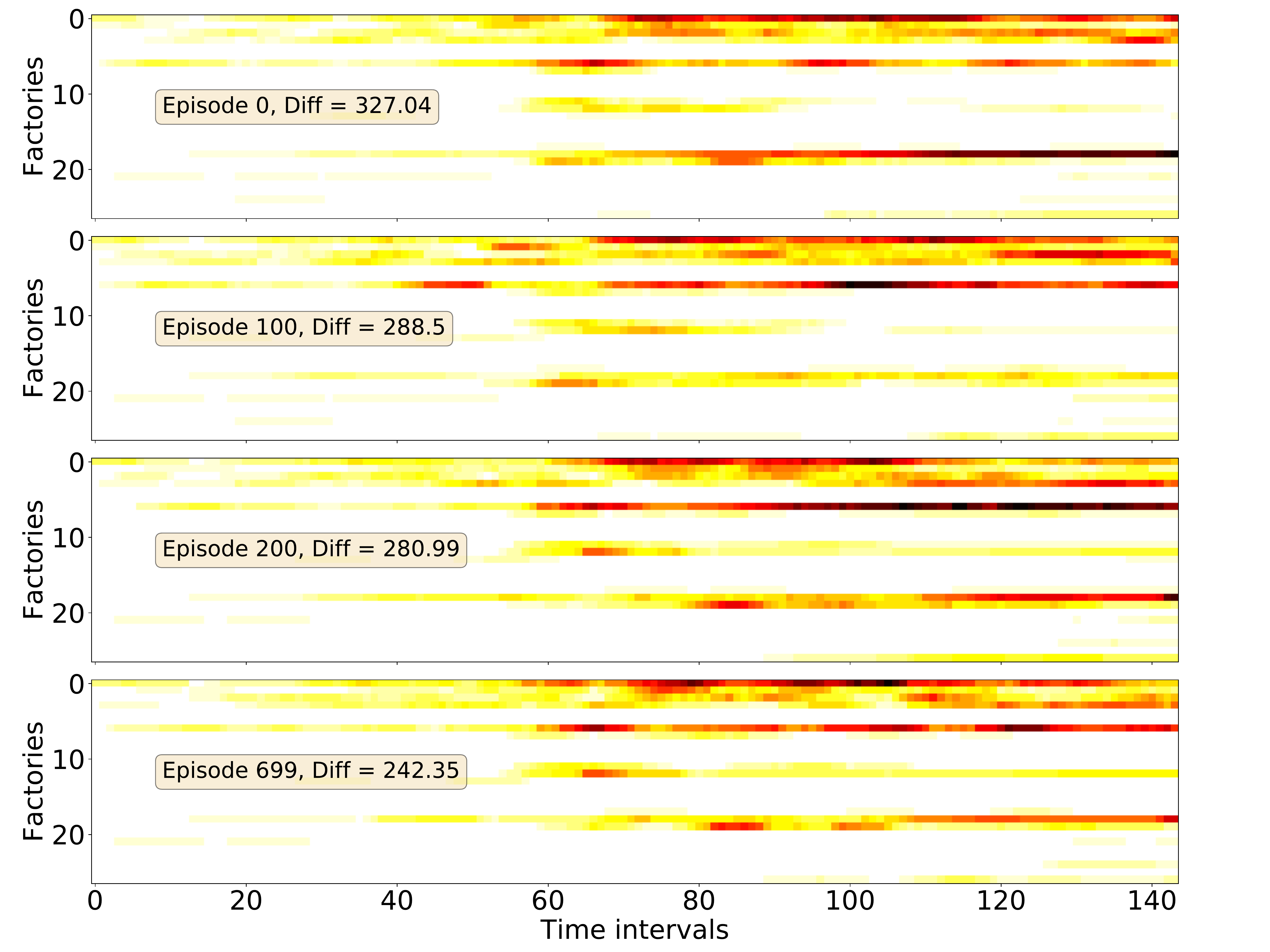}
		\label{fig: DGN iteration}
	}
	\subfigure[ST-DDGN]{
		\includegraphics[width=0.42\linewidth]{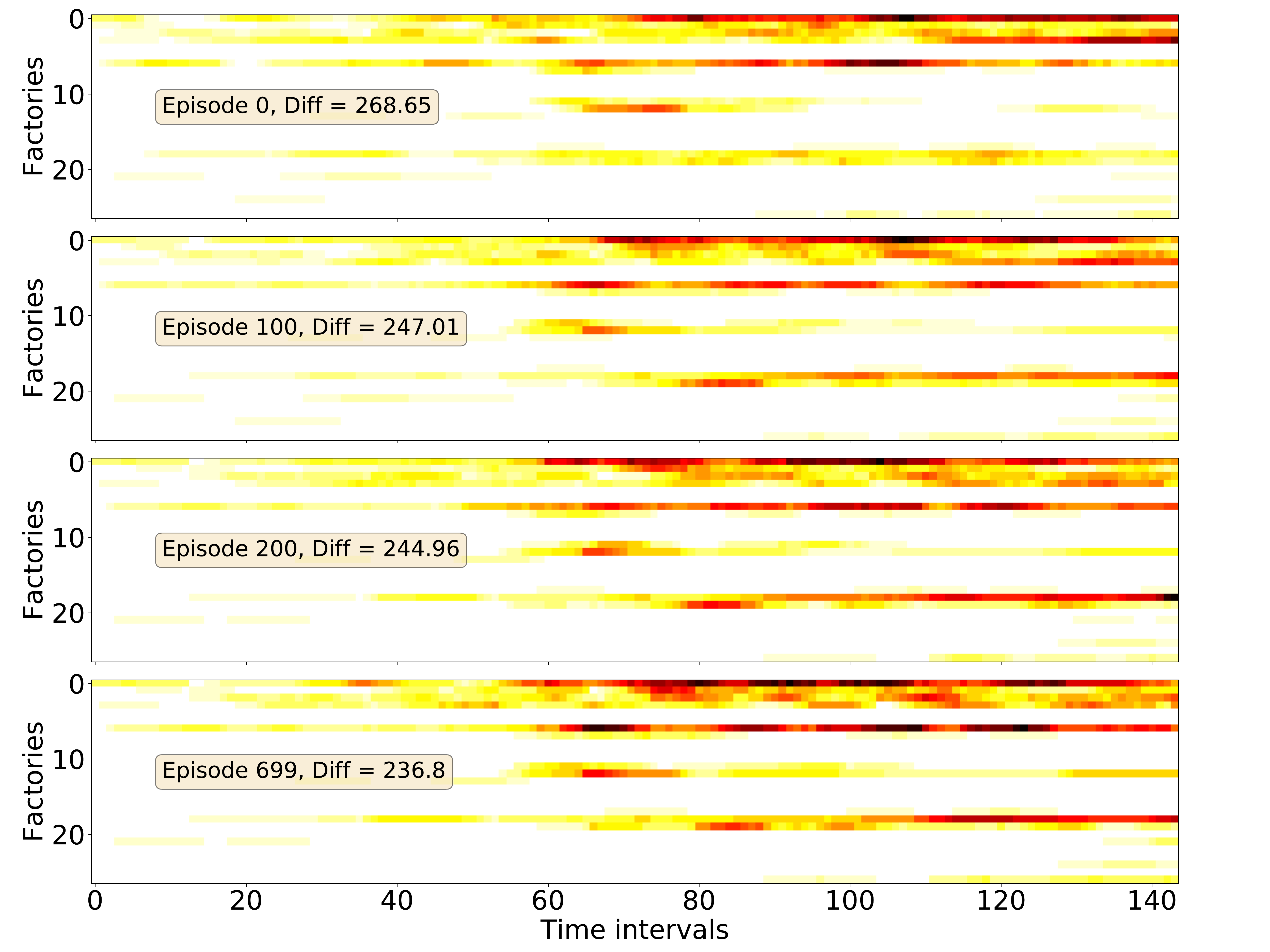}
		\label{fig: ST-DDGN iteration}
	}
	\caption{Spatial-temporal vehicle capacity distribution in policy learning episodes using various DRL methods. The spatial-temporal distribution of delivery capacity is similar to that of delivery orders, which consists of $27 \times 144$ grids. Differently, each grid represents how much the delivery capacity it is assigned. The darker a grid is, the larger delivery capacity the grid is assigned.}
	\label{fig:policy iteration}
\end{figure*}

\begin{figure}[h]
	\centering
	\includegraphics[width=0.9\linewidth]{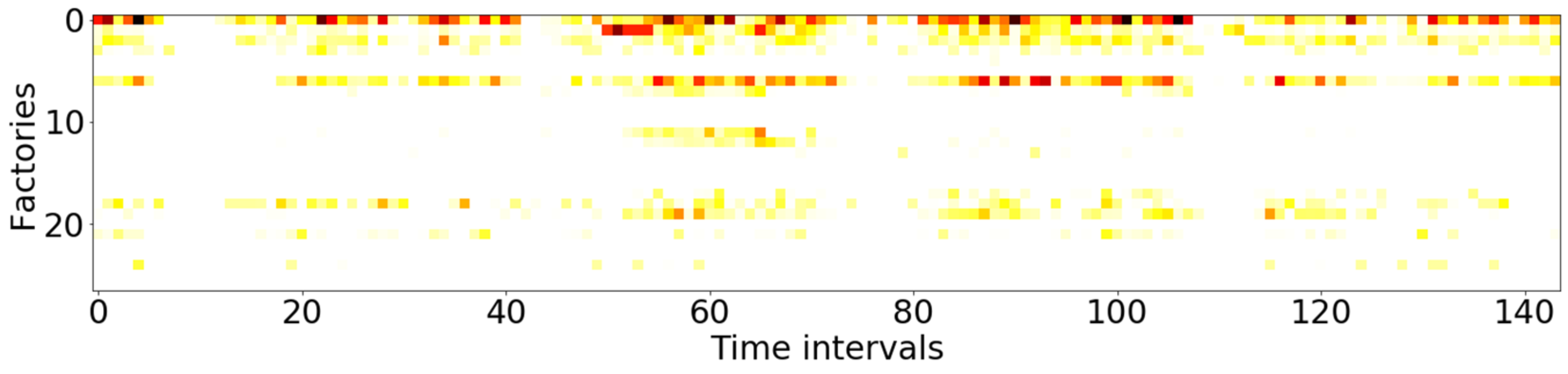}
	\caption{Spatial-temporal distribution of demand on large-scale instance}
	\label{fig: st_order_11}
\end{figure}

The convergence results of above four DRL methods are presented in Fig.~\ref{fig:Sequential}. With respect to NUV (see Fig.~\ref{fig:nuv-150}), all DRL methods learn a policy that can use less vehicles to serve all the orders than the baseline. Furthermore, it can be seen that models with graph convolution (DDGN and ST-DDGN) can learn better policies than DDQN and ST-DDQN. First, the training of DDGN and ST-DDGN converges much faster than that of the other two DRL methods. With regard to TC (see Fig.~\ref{fig:moc-150}), DDGN and ST-DDGN also perform much better than their counterparts. Although DDQN and ST-DDQN use the same number of vehicles with DDGN and ST-DDGN when converged, the total cost of DDQN and ST-DDQN is $5\%$ higher than that of DDGN and ST-DDGN. Besides, the learning curves in Fig.~\ref{fig:moc-150} show that ST-DDGN starts to converge at the $150$th episode while DDGN starts to converge at $200$th episode. Similar trend is also revealed on the learning curves of ST-DDQN and DDQN. Thus, it could be inferred that the introduction of ST Score indeed accelerates the policy learning. Because with the guide of ST Score, the neural network is supposed to take into consideration spatial-temporal balance between `supply and demand' when dispatching vehicles.

%

\subsubsection{Spatial-temporal learning in policy iteration}

In Section~\ref{sec: ablation studies}, we have inferred that the introduction of ST Score accelerates the policy learning. To support the inference, some evidences are given here.

The spatial-temporal distribution of delivery demand on a large-scale instance is presented in Fig.~\ref{fig: st_order_11}, which reveals the `hot spot' of delivery demand. We then train four DRL models (ST-DDGN, DGN, DQN and AC) and adopt the policies being optimized on the instance respectively. For each model, we record the spatial-temporal distribution of vehicle capacity at each episode. The results are presented in Fig.~\ref{fig:policy iteration}. Furthermore, we calculate the \textit{Frobenius} norm between demand distribution and capacity distribution at each episode, which is the meaning of `Diff' in each subfigure. We hope that the Diff decreases gradually when policy iterating. The results show that all the four DRL models indeed learned the spatial-temporal demand distribution when policy converged. Because in each subfigure of Fig.~\ref{fig:policy iteration}, the Diff decreases when the policy iterates. Furthermore, it is observable that 1) the Diff of ST-DDGN is the smallest $(236.8)$ among that of four DRL methods when policy converged, and 2) the Diff of ST-DDGN drops fastest among four DRL methods. To some extent, it can explain why ST-DDGN can learn faster and better than other DRL methods on large-scale instances.

To summarize, we think our agent learned with ST-DDGN has the following preferences in dispatching. First, the agent dispatches as many vehicles to the high demand grid as possible. Because, to some extent, the spatial-temporal distribution of delivery capacity is consistent with that of delivery demand. In additions, it seems that the agent would consider the future delivery demand when dispatching vehicles for current order since some of delivery orders are postponed to be served together with future orders (i.e., hitchhiking). Thus, we believe that the introduction of spatial-temporal metric is helpful for the proposed relational learning framework to solve DPDP.

\section{Supplementary Materials}
\label{sec: supplementary materials}
\textcolor{black}{
Due to the limitation of space, we put part of contents in an external link, including parameter/hyperparameter settings for various DRL methods, mathematical formulation of pickup and delivery problems and performance comparison between two divergence metrics used in our solution. Readers of interest can directly access the  \href{https://xijun-album.oss-cn-hangzhou.aliyuncs.com/ICDE2021submissions/supplements.pdf}{link}.}
\section{Conclusion}

\label{sec: conclusion}

In this paper, we have proposed a graph-based deep RL approach to solve the industry-scale DPDP. Our extensive experiments have shown that the algorithm can significantly improve the current solutions in practice on both efficiency and effectiveness. The algorithm has been in offline testing and is expected to be integrated into Huawei logistics management systems. In the future, we plan to update the proposed algorithm into a distributed architecture to further improve its performance and robustness.

\section{Acknowledgement}
This work was supported in part by the Fundamental Research Funds for the Central Universities (WK3490000004) and Natural Science Foundation of China (61822604, 61836006, U19B2026).

\bibliographystyle{IEEEtran}
\bibliography{my_ref}

\begin{thebibliography}{10}
\providecommand{\url}[1]{#1}
\csname url@samestyle\endcsname
\providecommand{\newblock}{\relax}
\providecommand{\bibinfo}[2]{#2}
\providecommand{\BIBentrySTDinterwordspacing}{\spaceskip=0pt\relax}
\providecommand{\BIBentryALTinterwordstretchfactor}{4}
\providecommand{\BIBentryALTinterwordspacing}{\spaceskip=\fontdimen2\font plus
\BIBentryALTinterwordstretchfactor\fontdimen3\font minus
  \fontdimen4\font\relax}
\providecommand{\BIBforeignlanguage}[2]{{%
\expandafter\ifx\csname l@#1\endcsname\relax
\typeout{** WARNING: IEEEtran.bst: No hyphenation pattern has been}%
\typeout{** loaded for the language `#1'. Using the pattern for}%
\typeout{** the default language instead.}%
\else
\language=\csname l@#1\endcsname
\fi
#2}}
\providecommand{\BIBdecl}{\relax}
\BIBdecl

\bibitem{cordeau2008recent}
J.-F. Cordeau, G.~Laporte, and S.~Ropke, ``Recent models and algorithms for
  one-to-one pickup and delivery problems,'' in \emph{The vehicle routing
  problem: latest advances and new challenges}.\hskip 1em plus 0.5em minus
  0.4em\relax Springer, 2008, pp. 327--357.

\bibitem{lu2004exact}
Q.~Lu and M.~Dessouky, ``An exact algorithm for the multiple vehicle pickup and
  delivery problem,'' \emph{Transportation Science}, vol.~38, no.~4, pp.
  503--514, 2004.

\bibitem{mahmoudi2016finding}
M.~Mahmoudi and X.~Zhou, ``Finding optimal solutions for vehicle routing
  problem with pickup and delivery services with time windows: A dynamic
  programming approach based on state--space--time network representations,''
  \emph{Transportation Research Part B: Methodological}, pp. 19--42, 2016.

\bibitem{mitrovic2004waiting}
S.~Mittrovic-Minic, G.~Laporte \emph{et~al.}, ``Waiting strategies for the
  dynamic pickup and delivery problem with time windows,'' \emph{Transportation
  Research Part B: Methodological}, vol.~38, no.~7, pp. 635--655, 2004.

\bibitem{gendreau2006neighborhood}
M.~Gendreau, F.~Guertin, J.-Y. Potvin, and R.~S{\'e}guin, ``Neighborhood search
  heuristics for a dynamic vehicle dispatching problem with pick-ups and
  deliveries,'' \emph{Transportation Research Part C: Emerging Technologies},
  vol.~14, no.~3, pp. 157--174, 2006.

\bibitem{pureza2008waiting}
V.~Pureza and G.~Laporte, ``Waiting and buffering strategies for the dynamic
  pickup and delivery problem with time windows,'' \emph{INFOR: Information
  Systems and Operational Research}, vol.~46, no.~3, pp. 165--175, 2008.

\bibitem{savelsbergh1998drive}
M.~Savelsbergh and M.~Sol, ``Drive: Dynamic routing of independent vehicles,''
  \emph{Operations Research}, vol.~46, no.~4, pp. 474--490, 1998.

\bibitem{li2019cooperative}
X.~Li, J.~Zhang, J.~Bian, Y.~Tong, and T.-Y. Liu, ``A cooperative multi-agent
  reinforcement learning framework for resource balancing in complex logistics
  network,'' in \emph{Proceedings of the 18th International Conference on
  Autonomous Agents and MultiAgent Systems}.\hskip 1em plus 0.5em minus
  0.4em\relax International Foundation for Autonomous Agents and Multiagent
  Systems, 2019, pp. 980--988.

\bibitem{khalil2017learning}
E.~Khalil, H.~Dai, Y.~Zhang, B.~Dilkina, and L.~Song, ``Learning combinatorial
  optimization algorithms over graphs,'' in \emph{Advances in Neural
  Information Processing Systems}, 2017, pp. 6348--6358.

\bibitem{kool2018attention}
W.~Kool, H.~van Hoof, and M.~Welling, ``Attention, learn to solve routing
  problems!'' \emph{arXiv preprint arXiv:1803.08475}, 2018.

\bibitem{balaji2019orl}
B.~Balaji, J.~Bell-Masterson, E.~Bilgin, A.~Damianou, P.~M. Garcia, A.~Jain,
  R.~Luo, A.~Maggiar, B.~Narayanaswamy, and C.~Ye, ``Orl: Reinforcement
  learning benchmarks for online stochastic optimization problems,''
  \emph{arXiv preprint arXiv}, 2019.

\bibitem{berbeglia2010dynamic}
G.~Berbeglia, J.-F. Cordeau, and G.~Laporte, ``Dynamic pickup and delivery
  problems,'' \emph{European journal of operational research}, vol. 202, no.~1,
  pp. 8--15, 2010.

\bibitem{saez2008hybrid}
D.~S{\'a}ez, C.~E. Cort{\'e}s, and A.~N{\'u}{\~n}ez, ``Hybrid adaptive
  predictive control for the multi-vehicle dynamic pick-up and delivery problem
  based on genetic algorithms and fuzzy clustering,'' \emph{Computers \&
  Operations Research}, vol.~35, no.~11, pp. 3412--3438, 2008.

\bibitem{zambaldi2018relational}
V.~Zambaldi, D.~Raposo, A.~Santoro, V.~Bapst, Y.~Li, I.~Babuschkin, K.~Tuyls,
  D.~Reichert, T.~Lillicrap, E.~Lockhart \emph{et~al.}, ``Relational deep
  reinforcement learning,'' \emph{arXiv preprint arXiv:1806.01830}, 2018.

\bibitem{sukhbaatar2016learning}
S.~Sukhbaatar, R.~Fergus \emph{et~al.}, ``Learning multiagent communication
  with backpropagation,'' in \emph{Advances in neural information processing
  systems}, 2016, pp. 2244--2252.

\bibitem{peng2017multiagent}
P.~Peng, Y.~Wen, Y.~Yang, Q.~Yuan, Z.~Tang, H.~Long, and J.~Wang, ``Multiagent
  bidirectionally-coordinated nets: Emergence of human-level coordination in
  learning to play starcraft combat games,'' \emph{arXiv preprint
  arXiv:1703.10069}, 2017.

\bibitem{yang2018mean}
Y.~Yang, R.~Luo, M.~Li, M.~Zhou, W.~Zhang, and J.~Wang, ``Mean field
  multi-agent reinforcement learning,'' \emph{arXiv preprint arXiv:1802.05438},
  2018.

\bibitem{jaques2018social}
N.~Jaques, A.~Lazaridou, E.~Hughes, C.~Gulcehre, P.~A. Ortega, D.~Strouse,
  J.~Z. Leibo, and N.~De~Freitas, ``Social influence as intrinsic motivation
  for multi-agent deep reinforcement learning,'' \emph{arXiv preprint
  arXiv:1810.08647}, 2018.

\bibitem{niepert2016learning}
M.~Niepert, M.~Ahmed, and K.~Kutzkov, ``Learning convolutional neural networks
  for graphs,'' in \emph{International conference on machine learning}, 2016,
  pp. 2014--2023.

\bibitem{jiang2018graph}
J.~Jiang, C.~Dun, and Z.~Lu, ``Graph convolutional reinforcement learning for
  multi-agent cooperation,'' \emph{arXiv preprint arXiv:1810.09202}, vol.~2,
  no.~3, 2018.

\bibitem{malysheva2018deep}
A.~Malysheva, T.~T. Sung, C.-B. Sohn, D.~Kudenko, and A.~Shpilman, ``Deep
  multi-agent reinforcement learning with relevance graphs,'' \emph{arXiv
  preprint arXiv:1811.12557}, 2018.

\bibitem{li2019towards}
R.~Li, A.~Jabri, T.~Darrell, and P.~Agrawal, ``Towards practical multi-object
  manipulation using relational reinforcement learning,'' \emph{arXiv preprint
  arXiv:1912.11032}, 2019.

\bibitem{zhang2016dnn}
J.~Zhang, Y.~Zheng, D.~Qi, R.~Li, and X.~Yi, ``Dnn-based prediction model for
  spatio-temporal data,'' in \emph{Proceedings of the 24th ACM SIGSPATIAL
  International Conference on Advances in Geographic Information Systems},
  2016, pp. 1--4.

\bibitem{pan2019urban}
Z.~Pan, Y.~Liang, W.~Wang, Y.~Yu, Y.~Zheng, and J.~Zhang, ``Urban traffic
  prediction from spatio-temporal data using deep meta learning,'' in
  \emph{Proceedings of the 25th ACM SIGKDD International Conference on
  Knowledge Discovery \& Data Mining}, 2019, pp. 1720--1730.

\bibitem{wang2018deep}
Z.~Wang, Z.~Qin, X.~Tang, J.~Ye, and H.~Zhu, ``Deep reinforcement learning with
  knowledge transfer for online rides order dispatching,'' in \emph{2018 IEEE
  International Conference on Data Mining (ICDM)}.\hskip 1em plus 0.5em minus
  0.4em\relax IEEE, 2018, pp. 617--626.

\bibitem{zhang2017taxi}
L.~Zhang, T.~Hu, Y.~Min, G.~Wu, J.~Zhang, P.~Feng, P.~Gong, and J.~Ye, ``A taxi
  order dispatch model based on combinatorial optimization,'' in
  \emph{Proceedings of the 23rd ACM SIGKDD International Conference on
  Knowledge Discovery and Data Mining}.\hskip 1em plus 0.5em minus 0.4em\relax
  ACM, 2017, pp. 2151--2159.

\bibitem{mnih2016a3c}
V.~Mnih, A.~P. Badia, M.~Mirza, A.~Graves, T.~P. Lillicrap, T.~Harley,
  D.~Silver, and K.~Kavukcuoglu, ``Asynchronous methods for deep reinforcement
  learning.'' in \emph{Proceedings of the 33rd International Conference on
  Machine Learning (ICML2016)}, 2016.

\bibitem{2015Continuous}
T.~P. Lillicrap, J.~J. Hunt, A.~Pritzel, N.~Heess, and D.~Wierstra,
  ``Continuous control with deep reinforcement learning,'' in \emph{Proceedings
  of the 4th International Conference on Learning Representations (ICLR2016)},
  2016.

\bibitem{lin2018efficient}
K.~Lin, R.~Zhao, Z.~Xu, and J.~Zhou, ``Efficient large-scale fleet management
  via multi-agent deep reinforcement learning,'' in \emph{Proceedings of the
  24th ACM SIGKDD International Conference on Knowledge Discovery \& Data
  Mining}, 2018, pp. 1774--1783.

\bibitem{vaswani2017attention}
A.~Vaswani, N.~Shazeer, N.~Parmar, J.~Uszkoreit, L.~Jones, A.~N. Gomez,
  {\L}.~Kaiser, and I.~Polosukhin, ``Attention is all you need,'' in
  \emph{Advances in neural information processing systems}, 2017, pp.
  5998--6008.

\bibitem{huang2017densely}
G.~Huang, Z.~Liu, L.~Van Der~Maaten, and K.~Q. Weinberger, ``Densely connected
  convolutional networks,'' in \emph{Proceedings of the IEEE conference on
  computer vision and pattern recognition}, 2017, pp. 4700--4708.

\bibitem{Grandinetti2014the}
L.~Grandinetti, F.~Guerriero, F.~Pezzella, and O.~Pisacane, ``The
  multi-objective multi-vehicle pickup and delivery problem with time
  windows,'' \emph{Procedia - Social and Behavioral Sciences}, vol. 111, pp.
  203--212, 2014.

\bibitem{mnih2015human}
V.~Mnih, K.~Kavukcuoglu, D.~Silver, A.~A. Rusu, J.~Veness, M.~G. Bellemare,
  A.~Graves, M.~Riedmiller, A.~K. Fidjeland, G.~Ostrovski \emph{et~al.},
  ``Human-level control through deep reinforcement learning,'' \emph{Nature},
  vol. 518, no. 7540, pp. 529--533, 2015.

\bibitem{ac2012}
T.~Degris, M.~White, and R.~Sutton, ``Off-policy actor-critic,'' in
  \emph{Proceedings of the 29th International Conference on Machine Learning,
  ICML 2012}, vol.~1, May 2012.

\end{thebibliography}

\begin{appendices}

\clearpage
\section{Algorithm pseudo code}
\label{sec: alogrithm pseudocode}
\small{
\begin{algorithm}[h]
	
	\caption{Simulator}
	\begin{algorithmic}[1] 
	
		\STATE \textbf{Input:} Historical order data and vehicle information
		\STATE \textbf{Initialization:} Initialize orders' and vehicles' state
		\FOR{each time interval $t$ in $episode$}
		\STATE \textbf{Initialize set of unassigned orders:} $O_t = \{o^i | t^i_c \in t\}$, and sort these orders in the ascending order of $t^i_c$
		\STATE \textbf{Update vehicles' real-time info:} current location, time interval $t$, route, \textit{etc.}
		\FOR{each unassigned order $o_t^i$ in $O_t$}
		\STATE \textbf{Implement learned policy:} 1) Get joint state $S_t^i$ of vehicles (i.e., the fleet) with respect to order $o_t^i$ via Algorithm~\ref{alg: route planner}; 2) Get the action $a^i_t$ from the policy and execute the action, i.e., dispatching a determined vehicle $k$ to serve order $o_t^i$
        \STATE \textbf{Update the selected vehicle's information:} $rp_k \leftarrow rp^i_{t,k}$, $d_{t,k} \leftarrow d^i_{t,k}$ using information from $S_t^i$
        
		\ENDFOR
		\ENDFOR
	\end{algorithmic}
	\label{alg: simulator}
\end{algorithm}

\begin{algorithm}[h]
	\caption{Route planner}
	\begin{algorithmic}[1]

		\STATE \textbf{Input:} Information of order $o_t^i$ and vehicle $k$
		\STATE Construct all possible temporary routes $RP_{t,k}^i=\{\tilde{rp}_{t,k}^i\}$ via inserting the pickup and delivery node of order $o_t^i$ into vehicle $k$'s current route $rp_k$ in an enumeration way
		\STATE Compute length $\tilde{d}^i_{t,k}$ of each temporary route $\tilde{rp}_{t,k}^i$
		\STATE Check each temporary route $\tilde{rp}_{t,k}^i$ to see whether it satisfies the time window, capacity and LIFO constraints
		\IF {there is at least one temporary route that satisfies all above three constraints}
		\STATE Set feasibility flag of vehicle $k$ $\mathit{w.r.t.}$ $o_t^i$: $fe^i_{t,k} \leftarrow 1$
		\STATE Set used flag $f_{t,k} \leftarrow 1 $ if vehicle $k$ has served orders before, else set $f_{t,k} \leftarrow 0 $
		\STATE Get the current length $d_{t,k}$ of route $rp_k$
		\STATE Get the shortest among those feasible temporary routes:
		$rp^{i}_{t,k}=\underset{\tilde{d}^i_{t,k}}{\operatorname{argmin}}\ \ \tilde{rp}_{t,k}^i$ and denote its length as $d_{t,k}^{i}$

		\STATE Calculate ST Score $\xi_{t,k}^i$ of the selected route $rp_{t,k}^i$
		\ELSE
		\STATE Set $fe^i_{t,k}  \leftarrow 0$,  $f_{t,k}  \leftarrow -1$, $d_{t,k}  \leftarrow -1$, $d_{t,k}^{i} \leftarrow -1$, and $\xi^i_{t,k}  \leftarrow -1$, $rp_{t,k}^i \leftarrow \{\}$
		\ENDIF
		\STATE \textbf{Output:}  $fe^i_{t,k}, d_{t, k}, d^i_{t,k}, \xi^i_{t,k}, f_{t,k}, t, rp_{t,k}^i$
	\end{algorithmic}
	\label{alg: route planner}
\end{algorithm}

\begin{algorithm}[h]
	\caption{Train ST-DDGN}
	\begin{algorithmic}[1] 
	
	    \STATE \textbf{Input:} the updating period $\mathcal{T}$ and $maxEpisode$
		\STATE \textbf{Initialize:} the policy $\pi$ and Q-function $Q$ with random parameters $\theta$ and the memory replay buffer $D=\{\}$
		\FOR{$episode=1$ to $maxEpisode$}
		\STATE Reset the environment in simulator
		\FOR{ $t=1$ to $T$}
		\FOR{each order $o_t^i$ in time interval $t$}
		\FOR{each vehicle $k=1$ to $K$}
		\STATE Get individual state $s_{t,k}^i$ via Algorithm~\ref{alg: route planner}
		\ENDFOR
		\STATE Construct the joint state $S_t^i$ via concatenating all vehicles' state $s_{t,k}^i$
		\STATE Sample an action with $\epsilon$-greedy policy: $a^i_t \sim \pi (S_t^i; \theta)$
		\STATE Compute the instant reward $r^i_t$ using Eq.(\ref{eq: instant reward})
        \IF{$o_t^i$ is the last order in time interval $t$}
        \STATE $\mathcal{F}^i_{t}\leftarrow \mathrm{True}$
        \ELSE
        \STATE $\mathcal{F}^i_{t}\leftarrow \mathrm{False}$
        \ENDIF
		\ENDFOR
		\ENDFOR
		\STATE Compute the long-term reward $\bar{r}$ according to Eq.(\ref{eq: long term reward})
		\STATE Compute the final reward $R^i_t$ according to Eq.(\ref{eq: final reward})
		\STATE Store all state transitions $\mathcal{S}=\{(S^{i}_t, a^i_t, \mathcal{F}^i_t, R^i_t, S^{i^\prime}_{t^\prime})\}$ during the $episode$ into $D$
		\STATE Sample a mini-batch $\mathcal{B}$ from the memory replay $D$
        \FOR{each state transition $(S^{i}_t, a^i_t, \mathcal{F}^i_t, R^i_t, S^{i^\prime}_{t^\prime})$ in $\mathcal{B}$}
		\IF{$\mathcal{F}^i_t$ is $\mathrm{True}$}
		\STATE $y^i_t\leftarrow R^i_t$
		\ELSE
		\STATE $y^i_t\leftarrow R^i_t + \gamma{Q'(S^{i^\prime}_{t^\prime}, \arg\max_{a^{i^\prime}_{t^\prime} \in\mathcal{A}} Q(S^{i^\prime}_{t^\prime},a^{i^\prime}_{t^\prime};\theta);\theta')} $
		\ENDIF
        \ENDFOR
        \STATE Calculate the loss $\mathcal{L}(\theta)=\frac{1}{|\mathcal{B}|}\sum^{}_{\mathcal{B}}(Q(S^i_t, a^i_t;\theta)-y^i_t)^2$
		\STATE Update the evaluate network $\theta \leftarrow \theta + \nabla_\theta\mathcal{L}(\theta)$
        \IF{$episode \% \mathcal{T} = 0$}
        \STATE Update the target network: $\theta' \leftarrow \theta$
		\ENDIF
        \ENDFOR
		
	\end{algorithmic}
	\label{alg: ddgn}
\end{algorithm}
}

\end{appendices}

\end{document}